\documentclass{article}
\usepackage{microtype}
\usepackage{graphicx}
\usepackage{subfigure}
\usepackage{booktabs} 
\usepackage{times}
\usepackage{placeins}
\usepackage{graphicx} 
\usepackage{subfigure} 

\usepackage{times}  
\usepackage{helvet} 
\usepackage{courier}  
\usepackage{url} 
\usepackage{graphicx}
\usepackage{datatool}
\usepackage{bbm}
\usepackage{algorithm}
\usepackage{algorithmic}
\usepackage{amsthm,amsmath,amssymb}
\usepackage{array}
\usepackage{enumerate}
\usepackage{multirow}
\theoremstyle{plain}
\newtheorem{theorem}{Theorem}
\newtheorem{definition}[theorem]{Definition}
\newtheorem{proposition}[theorem]{Proposition}
\newtheorem{corollary}[theorem]{Corollary}
\newtheorem{lemma}[theorem]{Lemma}

\DeclareMathOperator*{\argmin}{arg\,min}
\DeclareMathOperator*{\argmax}{arg\,max}

\newcommand{\feature}{\boldsymbol{x}}

\newcommand{\R}{\mathbb{R}}
\newcommand{\E}{\mathbb{E}}
\newcommand{\EJoint}{\mathop{\mathbb{E}}\limits_{(\feature,y) \sim D}}
\newcommand{\Epos}{\mathbb{E}_\mathrm{P}}
\newcommand{\Eneg}{\mathbb{E}_\mathrm{N}}

\newcommand{\zerooneloss}{\ell_{0\text{-}1}}

\newcolumntype{L}{>{$}l<{$}}
\newcolumntype{C}{>{$}c<{$}}
\usepackage{hyperref}

\usepackage[accepted]{icml2019}

\icmltitlerunning{On Symmetric Losses for Learning from Corrupted Labels}

\begin{document}

\twocolumn[
\icmltitle{On Symmetric Losses for Learning from Corrupted Labels}



\icmlsetsymbol{equal}{*}

\begin{icmlauthorlist}
\icmlauthor{Nontawat Charoenphakdee}{utokyo,riken}
\icmlauthor{Jongyeong Lee}{utokyo,riken}
\icmlauthor{Masashi Sugiyama}{riken,utokyo}
\end{icmlauthorlist}

\icmlaffiliation{utokyo}{Department of Computer Science, The University of Tokyo, Tokyo, Japan}
\icmlaffiliation{riken}{RIKEN Center of Artificial Intelligence Project, Tokyo, Japan}
\icmlcorrespondingauthor{Nontawat Charoenphakdee}{nontawat@ms.k.u-tokyo.ac.jp}
\icmlcorrespondingauthor{Jongyeong Lee}{lee@ms.k.u-tokyo.ac.jp}
\icmlcorrespondingauthor{Masashi Sugiyama}{sugi@k.u-tokyo.ac.jp}
\icmlkeywords{Machine Learning, ICML}
\vskip 0.3in
]



\printAffiliationsAndNotice{}  

\begin{abstract}
This paper aims to provide a better understanding of a symmetric loss. 
First, we emphasize that using a symmetric loss is advantageous in the balanced error rate (BER) minimization and area under the receiver operating characteristic curve (AUC) maximization from corrupted labels. 
Second, we prove general theoretical properties of symmetric losses, including a classification-calibration condition, excess risk bound, conditional risk minimizer, and AUC-consistency condition. 
Third, since all nonnegative symmetric losses are non-convex, we propose a convex barrier hinge loss that benefits significantly from the symmetric condition, although it is not symmetric everywhere. 
Finally, we conduct experiments to validate the relevance of the symmetric condition. 
\end{abstract} 

\section{Introduction}
In the real-world, it is unrealistic to expect that clean fully-supervised data can always be obtained. 
Weakly-supervised learning is a learning paradigm to mitigate this problem~\citep{zhou2017brief}.
For example, labelers are not necessarily experts or even human experts can make mistakes. 
Learning under noisy labels is an example of weakly-supervised learning that relaxes the assumption that labels are always accurate~\citep{noise1, noise2, noise3, natarajan2013learning}. Other examples of weakly-supervised learning are learning from positive and unlabeled data~\citep{du2015convex, du2014, kiryo2017}, learning from pairwise similarity and unlabeled data~\citep{baosu},  and learning from complementary labels~\citep{ishida}. 
\begin{figure}[t]
  \centering
\includegraphics[width=\columnwidth]{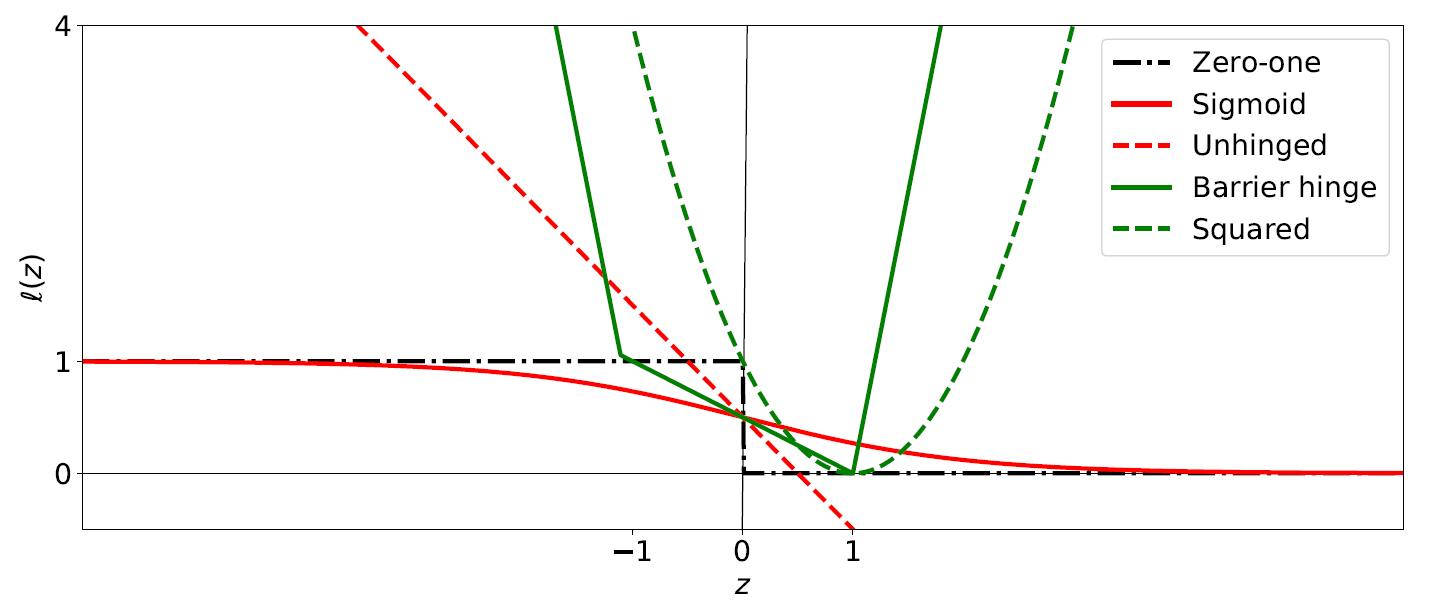}
\vspace{-0.3in}
  \caption{Examples of losses used in this paper. The zero-one loss, sigmoid loss, and unhinged loss are symmetric, i.e., $\ell(z)+\ell(-z)$ is a constant. The barrier hinge loss is our proposed loss.}
  \label{fig:all-loss-ex}
\end{figure}

A loss function that satisfies a symmetric condition has demonstrated its usefulness in weakly-supervised learning, e.g., one can use a symmetric loss to simplify a risk estimator in learning from positive-unlabeled data~\citep{du2014}. 
This simplification allows the use of a cost-sensitive learning library to implement the risk estimator directly. 
Not limited to the simplification of the risk estimator, symmetric losses are known to be robust in the symmetric label noise scenarios~\citep{manwani2013noise, ghosh2015making}.
However, the symmetric label noise assumption is restrictive and may not be practical since it assumes that a label of each pattern may flip independently with the same probability.

This paper elucidates the robustness of symmetric losses in a more general noise framework called the mutually contaminated distributions or corrupted labels framework~\citep{scott2013}.
Many weakly-supervised learning problems can be formulated in the corrupted labels framework~\citep{natarajan2013learning, lu2018minimal}.
Therefore, the robustness of learning from corrupted labels is highly desirable for many real-world applications.

Although it has been shown by~\citet{menon2015} that BER and AUC optimization from corrupted labels can be optimized without knowing the noise information, we point out that the use of non-symmetric losses may degrade the performance and therefore using a symmetric losses is preferable. 
Our experiments show that symmetric losses significantly outperformed many well-known non-symmetric losses when the given labels are corrupted.
Furthermore, we provide a better understanding of symmetric losses by elucidating several general theoretical properties of symmetric losses, including a classification-calibration condition, excess risk bound, conditional risk minimizer, and AUC-consistency.
We show that many well-known symmetric losses are suitable for both classification and bipartite ranking problems.
We also discuss the negative result of symmetric losses, which is the inability to recover the class probability given the risk minimizer.
This suggests a limitation to use such symmetric losses for a task that requires a prediction confidence such as learning with a reject option~\citep{chow1970, yuan2010}.

Unfortunately, it is known that a nonnegative symmetric loss must be non-convex~\citep{du2014, ghosh2015making}. 
~\citet{van2015learning} proposed an unhinged loss, which is convex, symmetric but negatively unbounded. 
In this paper, we propose a barrier hinge loss which is convex, nonnegative, and satisfies the symmetric condition in a subset of the domain space, not everywhere. 

\section{Preliminaries}
In this section, we review the notation and related work of symmetric losses and learning from corrupted labels.
\subsection{Notation}
 Let $\feature \in \R^d$ be a $d$-dimensional real-valued pattern, $y \in \{-1, +1\}$ denote a class label which can only be either positive or negative, and $g\colon\R^d \to \R$ denote a prediction function. 
In binary classification, we use sign$(g(\feature))$ to determine the predicted label of a prediction function, where sign$(g(\feature))=1$ if $g(\feature)>0$, $-1$ if $g(\feature)<0$, and $0$ otherwise.
$\Epos[\cdot] $ and $\Eneg[\cdot]$ denote the expectations of $\feature$ over $p(\feature|y=1)$ and $p(\feature|y=-1)$, respectively. 
$\eta(\feature)$ indicates the class probability $p(y=1|\feature)$ of a pattern $\feature$. 
In this paper, we consider a margin loss $\ell\colon \R \to \R$ that takes only one argument, which is typically~$yg(\feature)$. Table~\ref{table:optimal-loss} shows examples of margin losses.

\subsection{Symmetric Losses}
Note that the notion of a \emph{symmetric loss} can be ambiguous 
since there are many definitions of symmetric loss
(see ~\citet{natarajan2013learning, reid2010} for other definitions).
In this paper, we consider a \emph{symmetric} loss from the perspective that it is a margin loss, $\ell\colon \R \to \R$ that satisfies the symmetric condition, i.e., $\ell(z)+\ell(-z) = K$, where $K$ is a constant. Examples of such losses are the zero-one loss, unhinged loss and sigmoid loss which are described in Figure~\ref{fig:all-loss-ex}. 

The advantage of using a symmetric loss was investigated in the symmetric label noise scenario \citep{manwani2013noise, ghosh2015making, van2015learning}. 
The results from ~\citet{long2010random} suggested that convex losses are non-robust in this scenario and this motivated the use of a robust non-convex loss in the symmetric label noise scenario. 
\citet{ghosh2015making} proved that the symmetric condition is sufficient for a loss to be robust in this scenario. 
~\citet{van2015learning} later proposed an unhinged loss, which is the only possible convex loss to be symmetric, but it needs to be negatively unbounded. The negative unboundedness is not a common property for a loss function, which avoids the condition in~\citet{long2010random} to achieve the robustness in the symmetric label noise scenario. 
Another notable extension of a symmetric condition is the extension to a multiclass setting~\citep{ghosh2017robust}.

This paper considers a noise framework called mutually contaminated distributions or corrupted labels framework~\citep{scott2013}, where the symmetric label noise is a special case of the corrupted labels framework~\citep{menon2015}. Then, we discuss a problem of non-symmetric losses in this scenario and emphasize that advantage of symmetric losses. 

\subsection{Learning from Corrupted Labels}
In the corrupted labels scenario, we are given two sets of data drawn from the corrupted positive and corrupted negative marginal distributions respectively as follows:
\begin{align*}
\{\feature_i\}_{i=1}^{n_{\mathrm{CP}}} &\stackrel{\mathrm{i.i.d.}}{\sim} \pi p(\feature|y=1)+(1-\pi) p(\feature|y=-1) \text{,}\\
 \{\feature_j\}_{j=1}^{n_\mathrm{CN}}&\stackrel{\mathrm{i.i.d.}}{\sim} \pi' p(\feature|y=1)+(1-\pi') p(\feature|y=-1) \text{,}
\end{align*}
where $n_\mathrm{CP}$ denotes the number of corrupted positive patterns and $\pi$ is the class prior $p(y=1)$ for the corrupted positive distribution, i.e, a proportion of clean positive data in the corrupted positive data. $n_\mathrm{CN}$ and $\pi'$ are defined similarly for the corrupted negative data. We denote $X_\mathrm{CP}:=\{\feature_i\}_{i=1}^{n_{\mathrm{CP}}}$ as a corrupted positive sample and $X_\mathrm{CN}:=\{\feature_j\}_{j=1}^{n_{\mathrm{CN}}}$ as a corrupted negative sample. $p(\feature|y) $ denotes the class conditional density. In this setting, $\pi \neq \pi'$ but the class condition probabilities $p(\feature|y)$ are identical for both sets. Clean data implies $\pi = 1, \pi'=0$. The class prior in this case can also be interpreted as the noise rate \citep{menon2015}, where $(1-\pi)$ is the noise rate for positive data and $\pi'$ is the noise rate for negative data. We assume $\pi > \pi'$ for simplicity. Otherwise, labels from the classifier must be flipped.

\citet{menon2015} first showed that BER and AUC optimization from corrupted labels yield the same minimizer as minimizing from the clean labels. However, in this paper, we take a closer look of this problem and point out that the use of surrogate losses may yield different minimizers and degrade the performance.
Another notable work in this corrupted labels setting is the classification from two sets of unlabeled data~\citep{lu2018minimal}. They proposed an unbiased risk estimator for the classification error metric in this setting. BER is a special case of the classification error metric where the class prior is balanced. Nevertheless, their unbiased risk estimator requires the knowledge of the class priors of the two training distributions and the test distribution.
This paper only focuses on BER and AUC optimization and does not require any class prior information. 

\section{The Importance of Symmetric Losses in BER and AUC Optimization}
In this section, we show that using a symmetric loss is preferable for BER and AUC optimization from corrupted labels without class prior estimation. 
BER and AUC are popular metrics for imbalanced data classification~\citep{ber2, ber1}. Furthermore,
AUC is also known as an evaluation metric for bipartite ranking~\citep{narasimhan2013relationship, menon2016bipartite}. 
In the corrupted labels framework, the class prior estimation problem is known to be a bottleneck in this framework since it is an unidentifiable problem unless a restrictive condition is applied~\citep{blanchard2010semi, scott2015rate}. 
Thus, being able to minimize BER and AUC without estimating class priors is a great advantage in practice.

\noindent \textbf{Related work:}  \citet{menon2015} proved that for the zero-one loss, the clean and corrupted BER/AUC risks have the same minimizer. However, it remains unclear whether the same result holds for any surrogate losses. Later, \citet{van2015average} generalized the result of BER minimization in \citet{menon2015} from the zero-one loss to any symmetric losses.
In this paper, we analyze both BER and AUC optimization from corrupted labels by first proving the relationship between the clean surrogate risk and corrupted surrogate risk for \emph{any} surrogate losses. Our results indicate that using a non-symmetric loss may not yield the same minimizer for the clean and corrupted risks since it may suffer from excessive terms (see Sections~\ref{sec:auc-opt} and~\ref{sec:ber-opt}). Then, we clarify that similarly to BER minimization that was proven by~\citet{van2015average}, using a symmetric loss is also advantageous for AUC maximization. We are also the first to provide the experimental results for validating the advantage of symmetric losses for BER and AUC optimization from corrupted labels in practice.

\subsection{Area under the Receiver Operating Characteristic Curve (AUC) Maximization}\label{sec:auc-opt}

In AUC maximization, we consider the following AUC risk~\citep{narasimhan2013relationship}:
\begin{align} \label{aucrisk}
R^\ell_{\mathrm{AUC}}(g) = \Epos [ \Eneg[\ell(f(\feature_\mathrm{P}, \feature_\mathrm{N}))]] \text{,}
\end{align}
where  $f(\feature,\feature') = g(\feature)-g(\feature')$. 
The expected AUC score is $1-R^{\zerooneloss}_{\mathrm{AUC}}(g)$. Therefore, we can maximize the AUC score by minimizing the AUC risk. Since we do not have access to clean data, let us consider a corrupted AUC risk with a surrogate loss $\ell$ that treats $X_\mathrm{CP}$ as being positive and $X_{\mathrm{CN}}$ as being negative:
\begin{align*}
R^\ell_{\mathrm{AUC}\text{-}\mathrm{Corr}}(g) = \mathbb{E}_\mathrm{CP}[\mathbb{E}_{\mathrm{CN}}[\ell(f(\feature_\mathrm{CP},\feature_{\mathrm{CN}}))]]\text{.}
\end{align*}
The following theorem shows that by using a symmetric loss, the minimizers of $R^\ell_{\mathrm{AUC}\text{-}\mathrm{Corr}}(g)$ and $R^\ell_{\mathrm{AUC}}(g)$ are identical (its proof is given in Appendix).
\begin{theorem}
Let $\gamma^\ell(\feature,\feature') = \ell(f(\feature',\feature)) + \ell(f(\feature,\feature')) $. Then $R^\ell_{\mathrm{AUC}\text{-}\mathrm{Corr}}(g)$ can be expressed as
\begin{align*}
\begin{split}
R^\ell_{\mathrm{AUC}\text{-}\mathrm{Corr}}(g)  ={}& (\pi-\pi')R^\ell_{\mathrm{AUC}}(g) \\ &  + \underbrace{(1-\pi)\pi' \Epos[\Eneg[\gamma^\ell(\feature_\mathrm{P},\feature_\mathrm{N})]]}_\textup{Excessive term} \\ &+ \underbrace{\frac{\pi\pi'}{2} \E_{\mathrm{P'}}[\Epos[\gamma^\ell(\feature_{\mathrm{P'}},\feature_\mathrm{P})]]}_\textup{Excessive term} \\ & + \underbrace{\frac{(1-\pi)(1-\pi')}{2}   \E_{\mathrm{N'}}[\Eneg[\gamma^\ell(\feature_{\mathrm{N'}},\feature_\mathrm{N})]]}_\textup{Excessive term}\text{.}
\end{split}
\end{align*}
\end{theorem}
\begin{corollary}\label{auc-symmetric}
Let $\ell$ be a symmetric loss such that $\ell(z)+\ell(-z) = K$, where $K$ is a constant. $R^\ell_{\mathrm{AUC}\text{-}\mathrm{Corr}}(g)$ can be expressed as
\begin{align*}
R^\ell_{\mathrm{AUC}\text{-}\mathrm{Corr}}(g)  = (\pi-\pi')R^{\ell}_{\mathrm{AUC}}(g) +  K\left(\frac{1  -\pi + \pi'}{2}\right) \text{.}
\end{align*}
\end{corollary}

Corollary~\ref{auc-symmetric} can be obtained simply by substituting 
$\gamma^\ell(\feature,\feature')$ with $K$. 
This suggests that the excessive term becomes a constant when using a symmetric loss and guarantees that the minimizers of $R^\ell_{\mathrm{AUC}\text{-}\mathrm{Corr}}(g)$ and $R^\ell_{\mathrm{AUC}}(g)$ are identical. 
On the other hand, if a loss is non-symmetric, then the excessive terms are not constants and the minimizers of both risks may differ.
A special case of this setting where $\pi=1$ has been studied by~\citet{sakai2018}. They showed that a convex surrogate loss can be applied but $\pi'$ needs to be estimated in order to cancel the excessive term. 
By using a symmetric loss, the class prior estimation is not required and the given positive patterns can also be corrupted. 
More generally, our results indicate that using a symmetric loss for AUC maximization from corrupted labels yields the same minimizer as clean labels and can be applied to various weakly-supervised learning settings~\citep{natarajan2013learning,niu2016,baosu,lu2018minimal}. 

\subsection{Balanced Error Rate (BER) Minimization}\label{sec:ber-opt}
Consider the following misclassification risk:
\begin{align*}
R^\ell_{\mathrm{BER}}(g) &= \frac{1}{2} \left[\Epos\left[\ell(g(\feature))\right] + \Eneg\left[\ell(-g(\feature))\right] \right] \text{.}
\end{align*}
The BER minimization problem is equivalent to minimizing $R^{\zerooneloss}_{\mathrm{BER}}(g)$. i.e., the classification risk with the zero-one loss when the class prior of the test distribution is balanced.

Let us define
\begin{align*}
	R^\ell_{\mathrm{BER}\text{-}\mathrm{Corr}}(g) = \frac{1}{2} \big[ R_{\mathrm{CP}}^{\ell}(g) + R_{\mathrm{CN}}^{\ell}(g)  \big] \text{,}
\end{align*}
where 
\begin{align*}
R_{\mathrm{CP}}^{\ell}(g) &= \pi \Epos[\ell(g(\feature))]  + (1-\pi)\Eneg[\ell(g(\feature))] \text{,} \\
R_{\mathrm{CN}}^{\ell}(g) &= \pi' \Epos[\ell(-g(\feature))]  + (1-\pi')\Eneg[\ell(-g(\feature))]\text{.} 
\end{align*}
Then, we state the following theorem (its proof is given in Appendix).
\begin{theorem}
Let $\gamma^\ell(\feature) = \ell(g(\feature)) + \ell(-g(\feature))$, $R^\ell_{\mathrm{BER}\text{-}\mathrm{Corr}}(g)$ can be expressed as
\begin{align*}
R^\ell_{\mathrm{BER}\text{-}\mathrm{Corr}}(g) &= (\pi-\pi') {R^\ell_{\mathrm{BER}}(g)} \\& \quad + \underbrace{\frac{\pi' \Epos[\gamma^\ell(\feature)] + (1-\pi)\Eneg[\gamma^\ell(\feature)]}{2}}_\textup{Excessive term} \text{.}
\end{align*}
\end{theorem}
By observing an excessive term, we can directly obtain the following corollary, which coincides with the existing result by~\citet{van2015average}.
\begin{corollary}[\citet{van2015average}]\label{ber-symmetric}
Let $\ell$ be a symmetric loss such that $\ell(z)+\ell(-z) = K$, where $K$ is a constant. $R^\ell_{\mathrm{BER}\text{-}\mathrm{Corr}}(g)$ can be expressed as
\begin{align*}
    R^\ell_{\mathrm{BER}\text{-}\mathrm{Corr}}(g) = (\pi-\pi') R^{\ell}_{\mathrm{BER}}(g) + K\left(\frac{1-\pi+\pi'}{2}\right) \text{.}
\end{align*}
\end{corollary}
Similarly to Corollary~\ref{auc-symmetric}, if a loss $\ell$  is symmetric, then the excessive term is a constant and the minimizers of $R^\ell_{\mathrm{BER}\text{-}\mathrm{Corr}}(g)$ and $R^\ell_{\mathrm{BER}}(g)$ are guaranteed to be identical.
\begin{table*}[t]
\centering
\caption{Loss functions and their properties including the convexity, symmetricity, capability of recovering $\eta(\feature)$, and their conditional risk minimizers $f^{\ell*}(\feature)$. Although the conditional risk minimizers of each loss function are different, the sign of each minimizer sign$(f^{\ell*}(\feature))$ matches each other, which agrees with the Bayes-optimal classifier. The savage loss is proposed by~\citet{masnadi2009}. The minimizer $f^{\ell*}(\feature)$ of the ramp, sigmoid, and unhinged losses are unique if the prediction output is in $[-1,1]$.} \label{table:optimal-loss}
\begin{tabular}{|C|C|C | C | C | C | C|C|}
\hline
\text{Loss name} & \ell(z) &  f^{\ell*}(\feature) & \text{Convex} & \text{Symmetric} & \text{Recover $\eta(\feature)$}\\ \hline
\text{Zero-one} & -0.5 \mathrm{sign}(z) + 0.5&\mathrm{sign}(\eta(\feature)-0.5)&  \times &\checkmark&  \times\\ 
\text{Squared} & (1-z)^{2} &2\eta(\feature)-1&  \checkmark &\times&\checkmark\\ 
\text{Hinge} & \max(0, 1-z) &\mathrm{sign}(\eta(\feature)-0.5)&  \checkmark &\times &  \times\\ 
\text{Logistic} & \mathrm{log}(1+\exp(-z)) &\mathrm{log}\left(\frac{\eta(\feature)}{1-\eta(\feature)}\right)&  \checkmark &\times &\checkmark\\ 
\text{Savage}  & \left[(1+\exp(2z))^{2}\right]^{-1} &0.5\mathrm{log}\left(\frac{\eta(\feature)}{1-\eta(\feature)}\right)&\times &  \times &\checkmark\\ 
\text{Ramp}& \mathrm{max}(0, \mathrm{min}(1, 0.5-0.5z)) &\mathrm{sign}(\eta(\feature)-0.5)& \times &  \checkmark &  \times \\ 
 \text{Sigmoid} & \left[1+\exp(z)\right]^{-1} & \mathrm{sign}(\eta(\feature)-0.5) &\times &  \checkmark &  \times\\ 
  \text{Unhinged} & 1-z & \mathrm{sign}(\eta(\feature)-0.5) &\checkmark &  \checkmark &  \times\\ 
     \hline
\end{tabular}
\end{table*}
\section{Theoretical Properties of Symmetric Losses}
In this section, we investigate general theoretical properties of symmetric losses. 
Since all nonnegative symmetric losses are non-convex, many convenient conditions that assume a loss function is convex cannot be applied~\citep{zhang2004statistical, bartlett2006, gao2015consistency, niu2016}. Nevertheless, thanks to the symmetric condition, we show that it is possible to derive general theoretical properties of a symmetric loss. 

\subsection{Classification-calibration}
The main motivation to use a surrogate loss in binary classification is that the zero-one loss is discontinuous and therefore difficult to optimize~\citep{zeroonenphard2,zeroonenphard1}. 
A natural question is what kind of surrogate losses can be used instead of the zero-one loss. 
This problem has been studied extensively in binary classification~\citep{zhang2004statistical, bartlett2006}. 
Classification-calibration is known to be a minimal requirement of a loss function for the binary classification task (see ~\citet{bartlett2006} for more details on classification-calibration).

We derive the following theorem that establishes a necessary and sufficient condition for a symmetric loss to be classification-calibrated (its proof is given in Appendix).

\begin{theorem} \label{lem1} 
A symmetric loss $\ell\colon \R \to \R$ such that $\ell(z)+\ell(-z)$ is a constant is classification-calibrated if and only if $\inf\limits_{\alpha> 0} \ell(\alpha)< \inf\limits_{\alpha\leq 0} \ell(\alpha)$.
\end{theorem}
The following corollary is straightforward from the theorem above, but we emphasize it since it covers many surrogate symmetric losses, e.g., the sigmoid, ramp, and unhinged losses. 
\begin{corollary}\label{collem1} 
A non-increasing loss $\ell\colon \R \to \R$ such that $\ell(z)+\ell(-z)$ is a constant and $\ell'(0) < 0$, is classification-calibrated.
\end{corollary} 
Based on Theorem~\ref{lem1}, by simply checking the condition whether $\inf\limits_{\alpha> 0} \ell(\alpha)< \inf\limits_{\alpha\leq 0} \ell(\alpha)$ is necessary and sufficient to determine if a symmetric loss is classification-calibrated. 
Note that Corollary~\ref{collem1} is a sufficient condition that covers many symmetric losses such as the ramp loss and sigmoid loss. 
In general, the differentiability at zero of a symmetric loss is not required to verify the classification-calibrated condition unlike convex losses~\citep{bartlett2006}. 
Note that some specific symmetric losses such as the ramp loss and sigmoid loss were proven to be classification-calibrated~\citep{bartlett2006,niu2016}.
This paper provides a necessary and sufficient condition for all symmetric losses.

\subsection{Excess Risk Bound}
The excess risk bound provides a relationship between the excess risk of minimizing the misclassification risk with respect to the zero-one loss and the surrogate loss. It is known that an excess risk bound of a loss $\ell$ exists if and only if $\ell$ is classification-calibrated~\citep{bartlett2006}.
 
Consider the standard binary misclassification risk:
\begin{align}\label{pnrisk}
R^\ell(g) = \EJoint \left[ \ell(y g(\feature) )\right] \text{.}
\end{align}

The following theorem indicates an excess risk bound for any classification-calibrated symmetric loss (its proof is given in Appendix). 
\begin{theorem} \label{th:classification-calibration}
An excess risk bound of a classification-calibrated symmetric loss $\ell\colon \R \to \R$ such that $\ell(z)+\ell(-z)$ is a constant can be expressed as
\begin{align*}
R^{\zerooneloss}(g)-R^{\zerooneloss*} &\leq \frac{ R^{\ell}(g) - R^{\ell*}}{\inf\limits_{\alpha \leq 0}  \ell(\alpha) - \inf\limits_{\alpha > 0} \ell(\alpha)} \text{,}
\end{align*}
where $R^{\ell*} = \inf\limits_{g} R^{\ell}(g) $ and $R^{\zerooneloss*} = \inf\limits_{g} R^{\zerooneloss}(g)$.
\end{theorem}
The result suggests that the excess risk bound of any classification-calibrated symmetric loss is controlled only by the difference of the infima  $\inf\limits_{\alpha> 0} \ell(\alpha) - \inf\limits_{\alpha\leq 0} \ell(\alpha)$. 
Intuitively, the excess risk bound tells us that if the prediction function~$g$ minimizes the surrogate risk $R^{\ell}(g) = R^{\ell*}$, then the prediction function $g$ must also minimize the misclassification risk $R^{\zerooneloss}(g) = R^{\zerooneloss*}$.
\subsection{Inability to Recover the Class Probability $\eta(\feature)$}

We investigate the form of the conditional risk minimizer of a symmetric loss. The conditional risk minimizer is useful to know the behavior of a prediction function learned from minimizing such a surrogate loss. 
For example, we can recover a class probability $\eta(\feature)$ from a prediction function if a loss $\ell$ is a proper composite loss~\citep{buja2005loss,reid2010}. 
The mapping function to recover a class probability $\eta(\feature)$ depends on the conditional risk minimizer.
For example, one can recover the class probability $\eta(\feature)$ of the squared loss by the relationship $\eta(\feature) = \frac{ f^{\ell_{\mathrm{sq}}*}(\feature)+1}{2}$.
Table~\ref{table:optimal-loss} shows the examples of classification-calibrated losses and their conditional risk minimizers. 

Our following theorem states that the conditional risk minimizer of any classification-calibrated symmetric loss can be expressed as a scaled Bayes-optimal classifier (its proof is given in Appendix).
\begin{theorem} \label{condrisk}
Let $\ell$ be a symmetric loss $\ell\colon \R \to \R$ such that $\ell(z)+\ell(-z)$ is a constant and classification-calibrated, if the minimum of $\ell$ exists and $M \in \argmin\limits_{\alpha \in \R}\ell(\alpha)$. Then, the condition risk minimizer of $\ell$ can be expressed as follows:
\begin{align*}
f^{\ell*}(\feature) = M \, \mathrm{sign}(\eta(\feature)-\frac{1}{2}) \text{,}
\end{align*}
where $\eta(\feature) = p(y=1|\feature)$.
\end{theorem}
When a symmetric loss is classification-calibrated but the minimum does not exist, $M$ $\to$ $\infty$. Note that the minimizer of a symmetric loss does not need to be unique as there might exist many points that give the minimum value.

By observing the conditional risk minimizer in Theorem~\ref{condrisk}, it is obvious that the class probability $\eta(\feature)$ cannot be recovered from the conditional risk minimizer since it knows only whether $\eta(\feature) > \frac{1}{2}$. 
This similar property has been observed and well-studied for the hinge loss $\ell_{\mathrm{hinge}}(z)=\mathrm{max}(0, 1-z)$, where its minimizer is the Bayes-optimal classifier $\mathrm{sign}(\eta(\feature)-\frac{1}{2})$, which suggests that the hinge loss is not suitable for class probability estimation~\citep{ bartlett2007sparseness,buja2005loss,reid2010}. 

\subsection{AUC-consistency}
AUC-consistency is similar to classification-calibration but from the perspective of AUC maximization~\citep{gao2015consistency}, i.e., minimizing the pairwise conditional risk for AUC maximization instead of the pointwise conditional risk in bainry classification. 
The Bayes-optimal solution of AUC maximization is a function that has a strictly monotonic relationship with the class probability $\eta(\feature)$, which is a consequence of the Neyman-Pearson lemma~\citep{menon2016bipartite}.

Our following lemma states that classification-calibration is necessary for a symmetric loss to be AUC-consistent (its proof is given in Appendix).
\begin{lemma}
An AUC consistent symmetric loss $\ell\colon \R \to \R$ such that $\ell(z)+\ell(-z)$ is a constant, is classification-calibrated.
\end{lemma}

Next, an interesting question is whether all classification-calibrated symmetric losses are AUC-consistency. 
We prove by giving a counterexample that unfortunately this is not the case (its proof is given in Appendix).
\begin{proposition}\label{prop:gap}
Classification-calibration is necessary yet insufficient for a symmetric loss $\ell\colon \R \to \R$ such that $\ell(z)+\ell(-z)$ to be AUC-consistent.
\end{proposition}
Proposition~\ref{prop:gap} illustrates that there is a gap between classification-calibration and AUC-consistency for a symmetric loss. 
This gives rise to an important question whether well-known symmetric losses are AUC-consistent.
We elucidate the positive result by establishing a sufficient condition for a symmetric loss to be AUC-consistent, which covers almost all existing surrogate symmetric losses to the best of our knowledge (its proof is given in Appendix).
\begin{theorem}~\label{aucconsistent}
A non-increasing loss $\ell\colon \R \to \R$ such that $\ell(z)+\ell(-z)$ is a constant and $\ell'(0) < 0$, is AUC-consistent.
\end{theorem}

With Corollary~\ref{collem1}  and Theorem~\ref{aucconsistent}, we show that a non-increasing symmetric loss that $\ell'(0) < 0$ is sufficient to be both classification-calibrated and AUC-consistent. 
Such conditions are not difficult to satisfy in practice. 
In fact, most surrogate symmetric losses that we are aware of satisfy this condition. 
Thus, the choice of symmetric losses is highly flexible for both the classification and bipartite ranking problems.
\section{Barrier Hinge Loss}
\begin{figure}
  \centering
\includegraphics[width=\columnwidth]{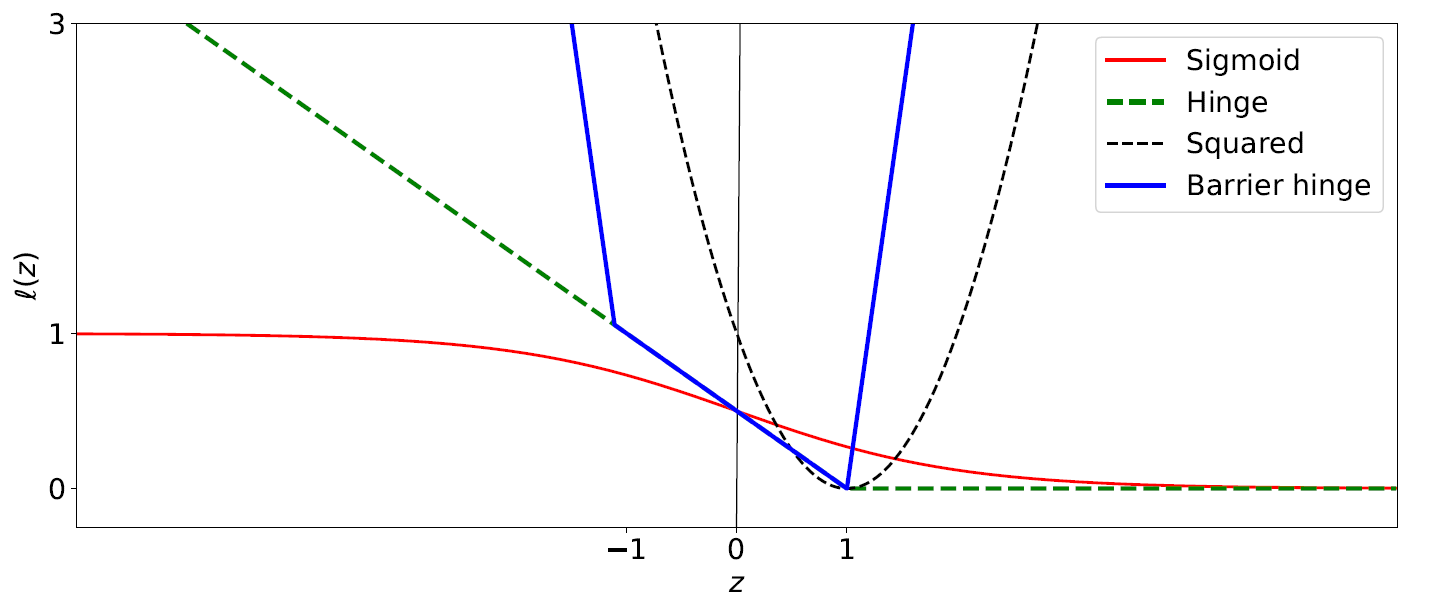}
\vspace{-0.35in}
  \caption{The barrier hinge loss scaled by $0.5$ with $b=10, r=1$: $\ell(z) = 0.5\mathrm{max}(-10(1+z)+1, \mathrm{max}(10(z-1), 1-z))$, the hinge loss: $\ell(z) = \mathrm{max}(0, 0.5-0.5z)$ and the sigmoid loss: $\ell(z) =\left[1+\exp(z)\right]^{-1}$. The symmetric property holds for the barrier hinge loss for $z \in [-1, 1]$.}
\label{fig:barrier-hinge}
\end{figure}
In this section, we propose a convex loss that benefits from the symmetric condition although it is not symmetric everywhere. Note that it is impossible to have a nonnegative symmetric loss~\citep{du2014,ghosh2015making}.
Our main idea to compensate this problem is to construct a loss that does not have to satisfy the symmetric condition everywhere, i.e ., $\ell(z)+\ell(-z)$ is a constant for every $z \in \R$. 
In this case, it is possible to find a classification-calibrated convex loss function that satisfies the symmetric condition only for an interval in $\R$. 
For example, the hinge loss satisfies the symmetric condition for $z \in [-1, 1]$. 
Nevertheless, the symmetric condition does not hold for $z$ when $z \notin [-1,1]$ and might suffer from the excessive term. 
Motivated by this observation, we propose a $\emph{barrier hinge loss}$, which is a loss that satisfies a symmetric condition not everywhere and gives a large penalty when $z$ is outside of the interval that is symmetric regardless of the correctness of the prediction.
\begin{figure}
  \centering
\includegraphics[width=\columnwidth]{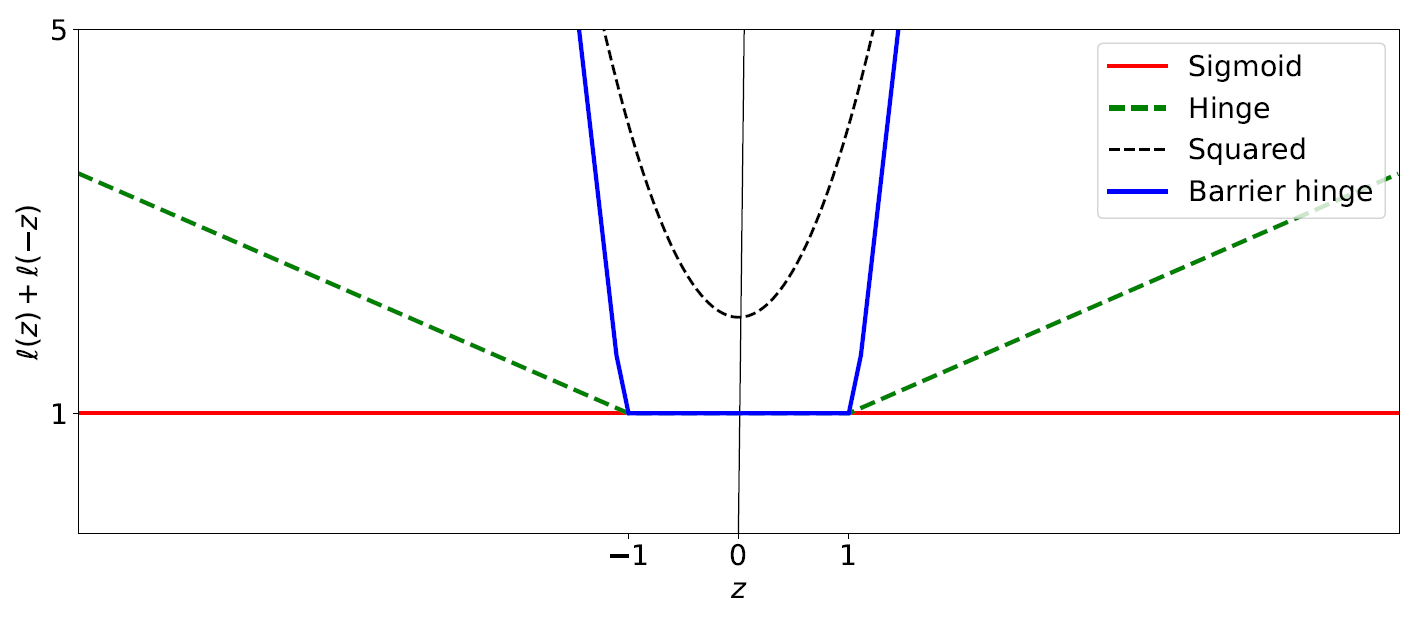}
\vspace{-0.35in}
  \caption{The plot of $\ell(z)+\ell(-z)$ of classification-calibrated losses. Only the sigmoid loss is symmetric. The hinge loss and barrier hinge loss satisfy the symmetric condition in $z \in [-1, 1]$.}
\label{fig:sym-visual}
\end{figure}
\begin{definition}
{\rm
A barrier hinge loss is defined as
\begin{align*}
\ell(z) = \mathrm{max}(-b(r+z)+r, \mathrm{max}(b(z-r), r-z)) \text{,}
\end{align*}
where $b > 1$ and $r>0$.}
\end{definition}

Figure~\ref{fig:barrier-hinge} shows a scaled barrier hinge loss with a specific parameter. 
Since a barrier hinge loss is convex, it is simple to verify that it is classification-calibrated since the derivative of the barrier hinge loss at zero is negative~\citep{bartlett2006}. 
Intuitively, barrier hinge losses are designed to give a very high penalty when $z$ is in the non-symmetric area. 
As a result, a prediction function which is learned from a barrier hinge loss has an incentive to give a prediction value inside the symmetric area.
The parameter $r$ determines the width of the region that satisfies the symmetric property while the parameter $b$ determines the slope of the penalty when $z$ is in the non-symmetric area ($b$ is expected to be a large value). 
In the experiment section, we show that our barrier hinge loss benefits from the symmetric condition and more robust than other non-symmetric losses. 
For fairness, we fix $b=200$ and $r=50$ for all datasets in the experiment section.
Hence, one can further tune the parameters $b$ and $r$ to achieve a more preferable performance.

It is important to note that if we restrict the output of a loss to be in a symmetric region, e.g., $g(\feature) \in [-1, 1]$ and $r \geq 1$, using the barrier hinge loss, unhinged loss, or standard hinge loss, are equivalent. Thus, the barrier hinge loss can also be viewed as a soft-constrained version of the unhinged loss.

\begin{figure*}
  \centering
  \vspace{-0.1in}
\includegraphics[width=\textwidth]{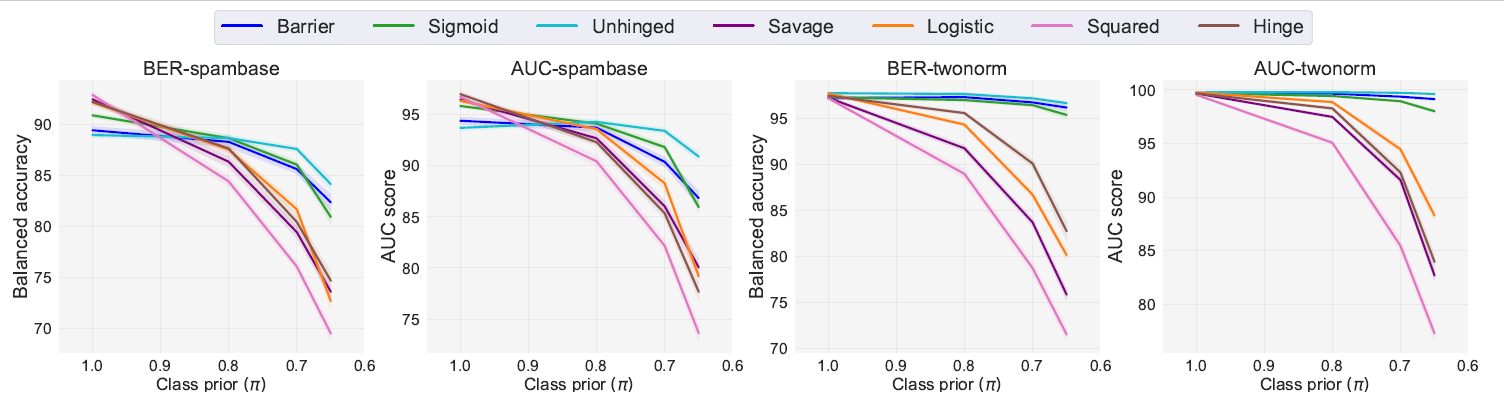}
\caption{Mean balanced accuracy (1-BER) and AUC score using multilayer perceptrons (rescaled to 0-100) with varying noise rates $(\pi=1.0, \pi'=0.0), (\pi=0.8, \pi'=0.3), (\pi=0.7, \pi'=0.4), (\pi=0.65, \pi'=0.45)$.  The experiments were conducted 20 times.}
\label{fig:uci}
\end{figure*}
\begin{table*}
\centering
\caption{Mean balanced accuracy (BAC=1-BER) and AUC score using multilayer perceptrons (rescaled to 0-100), where $\pi = 0.65$ and $\pi'=0.45$. Outperforming methods are highlighted in boldface using one-sided t-test with the significance level 5\%. The experiments were conducted 20 times.}
\label{table:worst-noise-uci-main-part}
\vspace{0.1in}
\begin{tabular}{|L|L|L|L | L | L | L|L|L|  }
\hline
\text{Dataset} & \text{Task} &\text{Barrier}& \text{Unhinged}& \text{Sigmoid}& \text{Logistic}& \text{Hinge}& \text{Squared}& \text{Savage}\\ \hline
\multirow{ 2}{*}{\text{spambase} }
	& \text{BAC} &82.3 (0.8) & \textbf{84.1 (0.6)} &80.9 (0.6) &72.6 (0.7) &74.7 (0.7) &69.5 (0.7) &73.6 (0.6)\\ 
	{} & \text{AUC} &86.8 (0.7) & \textbf{90.9 (0.4)} &86.0 (0.4) &79.2 (0.8) &77.7 (0.7) &73.6 (0.8) &80.1 (0.8)\\ 
 \hline
  \multirow{ 2}{*}{\text{waveform} }
 	& \text{BAC} & \textbf{86.1 (0.4)} & \textbf{87.1 (0.6)} &85.4 (0.6) &75.8 (0.7) &78.3 (0.7) &69.2 (0.6) &73.2 (0.6)\\ 
 	{} & \text{AUC}  & \textbf{92.2 (0.4)} & \textbf{91.7 (0.6)} & \textbf{90.9 (0.6)} &82.3 (0.7) &79.8 (0.9) &75.1 (0.7) &80.1 (0.6)\\ 
  \hline
  \multirow{ 2}{*}{\text{twonorm} }
 	& \text{BAC} & \textbf{96.2 (0.3)} & \textbf{96.7 (0.2)} &95.4 (0.4) &80.2 (0.5) &82.8 (0.9) &71.6 (0.7) &75.9 (0.6)\\ 
 	{} & \text{AUC} &99.1 (0.1) & \textbf{99.6 (0.0)} &98.0 (0.2) &88.3 (0.5) &83.9 (0.7) &77.3 (0.7) &82.7 (0.5)\\ 
  \hline
   \multirow{ 2}{*}{\text{mushroom} }
 	& \text{BAC} & \textbf{93.4 (0.8)} &91.1 (0.9) & \textbf{94.4 (0.7)} &81.3 (0.5) &84.5 (1.0) &72.2 (0.6) &79.5 (0.8)\\ 
 	{} & \text{AUC} & \textbf{98.4 (0.2)} &97.2 (0.4) & \textbf{97.8 (0.3)} &89.0 (0.5) &82.2 (0.6) &77.8 (0.6) &88.1 (0.7)\\ 
  \hline
\end{tabular}
\end{table*}
\section{Experimental Results}
In this section, we present experimental results of BER and AUC optimization from corrupted labels. 
We used the balanced accuracy (1-BER) to evaluate the performance of BER minimization and the AUC score for AUC maximization.
We also rescaled the score to be from 0 to 100.
Note that higher balanced accuracy and AUC score are better.
Training data were corrupted manually by simply mixing positive and negative data according to the class prior of the corrupted positive and corrupted negative data, i.e., $\pi$ and $\pi'$. 
We compare the following loss functions: the squared loss, logistic loss, exponential loss, hinge loss, savage loss, sigmoid loss, unhinged loss, and barrier loss.
Note that the class prior information is not given to the classifier.
Moreover, only the sigmoid loss and unhinged loss are symmetric while our proposed barrier loss is not symmetric everywhere but is designed to benefit from the symmetric condition. 
One might suspect that the improvement of the performance comes from the fact that these symmetric losses are bounded from above and therefore more robust against noise. To emphasize the importance of the symmetric property, we also compare the performance with the savage loss, a loss function which is bounded and has demonstrated its robustness against outliers in classification~\citep{masnadi2009}. We also found that the double hinge loss~\citep{du2015convex} performed similarly to the hinge loss and thus we omit the results.

We design the experiments to answer the following three questions. First, does the symmetric condition helps significantly in BER and AUC optimization from corrupted labels? 
Second, do we need a loss to be symmetric everywhere to benefit from the robustness of symmetric losses?
Third, does the negative unboundedness of the unhinged loss degrade the practical performance?

\subsection{Experiments on UCI and LIBSVM Datasets}
In this experiment, we used the one hidden layer multilayer perceptron $d-500-1$ as a model.
We used datasets from the UCI machine learning repository~\citep{dataset3} and LIBSVM~\citep{dataset2}. 
Training data consists of 500 corrupted positive data, 500 corrupted negative data, and balanced 500 clean test data.
More details on the implementation, datasets, and full experimental results using more datasets can be found in Appendix. 
The objective functions of the neural networks were optimized using AMSGRAD~\citep{adam_amsgrad}. The experiment code was implemented with Chainer~\citep{tokui2015chainer}. 

\begin{figure*}
  \centering
  \vspace{-0.1in}
\includegraphics[width=\textwidth]{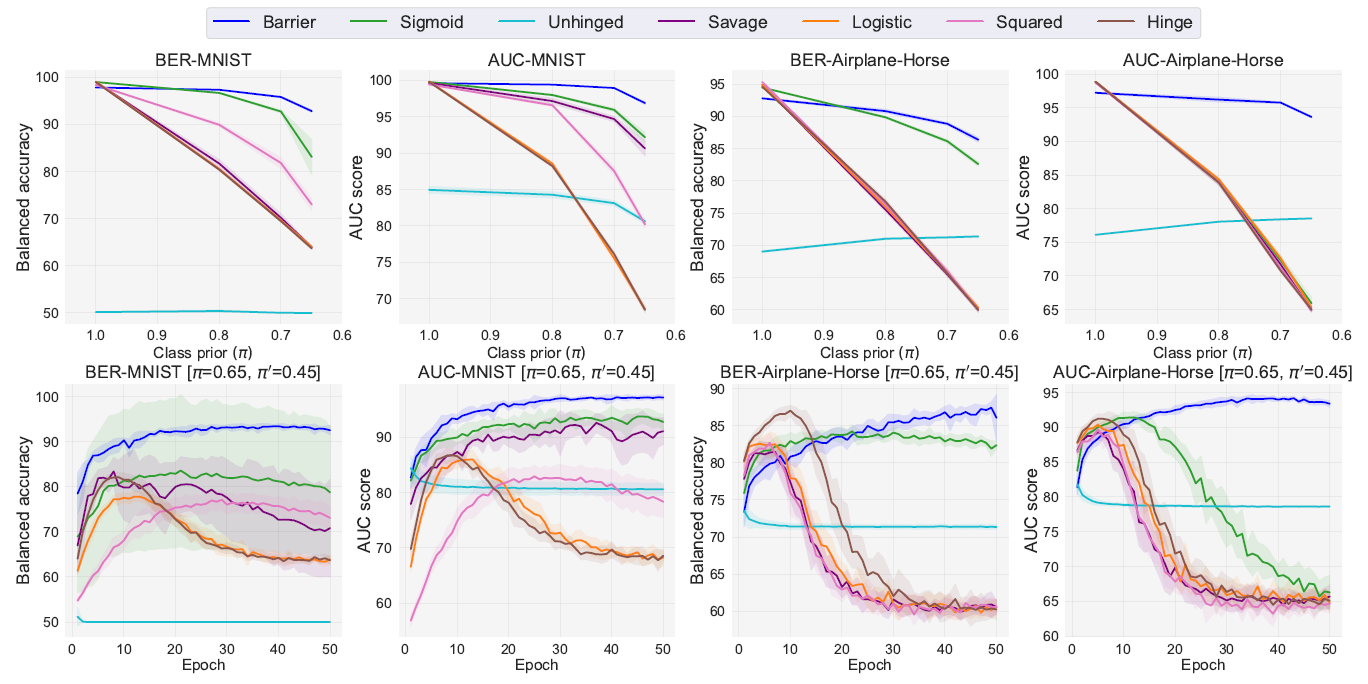}
\caption{Mean balanced accuracy (1-BER) and AUC score using convolutional neural networks (rescaled to 0-100).
(Top) the varying noise rates ranged from$(\pi=1.0, \pi'=0.0), (\pi=0.8, \pi'=0.3), (\pi=0.7, \pi'=0.4), (\pi=0.65, \pi'=0.45)$.
(Bottom) the noise rate is $\pi = 0.65$ and $\pi'=0.45$. The experiments were conducted 10 times.}
\label{fig:deep-main}
\end{figure*}

Figure \ref{fig:uci} shows the performance of BER and AUC optimization with varying noise rates $(\pi=1.0, \pi'=0.0), (\pi=0.8, \pi'=0.3), (\pi=0.7, \pi'=0.4), (\pi=0.65, \pi'=0.45)$.
Table \ref{table:worst-noise-uci-main-part} also shows the results where labels are highly corrupted ($\pi=0.65$ and $\pi'=0.45$).
Although the savage loss is a bounded loss, its performance is not desirable when the labels are corrupted. 
It can be observed that when the data is clean ($\pi=1.0$ and $\pi'=0.0$), the performance of all losses are not significantly different. 
However, as the noise rate increases, the sigmoid loss, unhinged loss, and barrier loss significantly outperform other losses in this experiment. 
This suggests that only using a bounded loss is not sufficient to perform BER minimization from corrupted labels effectively.
Therefore, the experimental results support our hypothesis that using symmetric losses can be preferable in the BER minimization problem from corrupted labels.

In this experiment, the unhinged loss performs well although it is negatively unbounded. This positive result of the unhinged loss agrees with \citet{van2015learning}, where they used a linear-in-input model.
However, our next experiment shows that the performance of the unhinged loss is less desirable when deeper neural networks are applied. 
\subsection{Experiments on MNIST and CIFAR-10}
In this experiment, we used MNIST~\citep{lecun1998mnist} (Odd vs~Even) and CIFAR-10 (Airplane vs~Horse)~\citep{cifar10} as the datasets. 
We used  the convolutional neural networks as the models for all losses. 
Full experimental results including the experiments on additional eight pairs of CIFAR10 and the implementation details can be found in Appendix.
The objective functions were optimized using AMSGRAD~\citep{adam_amsgrad}. The experiment code was implemented with PyTorch~\citep{pytorch2017}.

Figure \ref{fig:deep-main} (top) shows the performance on BER and AUC optimization with varying noise rates similarly to the previous experiment. 
It is observed that the unhinged loss failed miserably in BER minimization, although it outperformed other baselines when the labels are highly corrupted in CIFAR-10 (Airplane vs~Horse). 
Our proposed barrier hinge loss is observed to be advantageous in this experiment.

Figure \ref{fig:deep-main} (bottom) shows the performance on BER and AUC optimization from highly corrupted labels~$(\pi=0.65, \pi'=0.45)$ as the training epoch increases. The unhinged loss is observed to converge very quickly but its performance is marginal. The performance of the barrier hinge loss is preferable and does not degrade as the number of epoch increases. For the sigmoid loss, it is observed that the performance also degraded for the AUC maximization in CIFAR-10 as the epoch increases although it degraded slower than other losses that do not benefit from the symmetric condition.

In summary, our experimental results support that the symmetric condition significantly contributes to improving the performance on BER and AUC optimization from corrupted labels. Our barrier hinge loss, which is not symmetric everywhere, also demonstrated its robustness in this experiment. Finally, the unhinged loss is observed to perform poorly when complex models such as the convolutional neural networks are applied for which the potential reason can be the negative unboundedness of the unhinged loss. 

\section{Conclusion}
We analyze a class of symmetric losses. 
We showed that the symmetric condition of a loss contributes to the robustness of the BER and AUC optimization from corrupted labels. 
Moreover, we proved the general theoretical results to provide a better understanding of symmetric losses. 
We also proposed a convex barrier hinge loss that is not symmetric everywhere but benefits greatly from the symmetric condition.
The experimental results showed the advantage of using a symmetric loss for the BER and AUC optimization from corrupted labels and also illustrated the problem when a loss is negatively unbounded, such as the unhinged loss.

\section*{Acknowledgement}
We thank Han Bao and Zhenghang Cui for helpful discussion. We also thank anonymous reviewers for providing insightful comments. NC was supported by MEXT scholarship and MS was supported by JST CREST JPMJCR18A2.

\bibliographystyle{icml2019}

\appendix
\onecolumn
\section{Proofs}
We provide the proofs in this section.
\subsection{Proof of Theorem 1}
\begin{proof}  Recall that the AUC risk is:
\begin{align*}
R^\ell_{\mathrm{AUC}}(g) = \Epos [ \Eneg[\ell(f(\feature_\mathrm{P}, \feature_\mathrm{N}))]] \text{.}
\end{align*}
 Corrupted AUC risk where $X_\mathrm{CP}$ is assigned to be positive and $X_{\mathrm{CN}}$ as negative:
\begin{align*}
R^\ell_{\mathrm{AUC}\text{-}\mathrm{Corr}}(g) = \mathbb{E}_\mathrm{CP}[\mathbb{E}_{\mathrm{CN}}[\ell(f(\feature_\mathrm{CP},\feature_{\mathrm{CN}}))]]\text{.}
\end{align*}
where
\begin{align*}
R_{\mathrm{CP}}^{\ell}(g) &= \pi \Epos[\ell(g(\feature))]  + (1-\pi)\Eneg[\ell(g(\feature))]  \text{,}\\
R_{\mathrm{CN}}^{\ell}(g) &= \pi' \Epos[\ell(-g(\feature))]  + (1-\pi')\Eneg[\ell(-g(\feature))] \text{.}
\end{align*}
$R^\ell_{\mathrm{AUC}\text{-}\mathrm{Corr}}(g) $ can be rewritten as follows:
\begin{align*}
R^\ell_{\mathrm{AUC}\text{-}\mathrm{Corr}}(g) ={}& \pi' \mathbb{E}_{\mathrm{CP}}[\Epos[\ell(f(\feature_{\mathrm{CP}},\feature_\mathrm{P}))]] + (1-\pi')\mathbb{E}_{\mathrm{CP}}[\Eneg[\ell(f(\feature_{\mathrm{CP}},\feature_\mathrm{N}))]] ]\\
\begin{split}
    ={}& \pi \pi' \E_{\mathrm{P'}}[\Epos[\ell(f(\feature_{\mathrm{P'}},\feature_\mathrm{P}))]] + (1-\pi)\pi' \Eneg[\Epos[\ell(f(\feature_\mathrm{N},\feature_\mathrm{P}))]] \\ &+ \pi(1-\pi') \Epos[\Eneg[\ell(f(\feature_\mathrm{P},\feature_\mathrm{N}))]] \\ &+ (1-\pi)(1-\pi')\E_{\mathrm{N'}}[\Eneg[\ell(f(\feature_{\mathrm{N'}},\feature_\mathrm{N}))]]  \text{.}
\end{split}
\end{align*}
Let 
\begin{align*}
A &=  \E_{\mathrm{P'}}[\Epos[\ell(f(\feature_{\mathrm{P'}},\feature_\mathrm{P}))]] \, \text{,}\\
B &=  \Eneg[\Epos[\ell(f(\feature_\mathrm{N},\feature_\mathrm{P}))]]\text{,} \\
C &=  \Epos[\Eneg[\ell(f(\feature_\mathrm{P},\feature_\mathrm{N}))]]  = R^\ell_{\mathrm{AUC}}(g) \text{,} \\
D &=  \E_{\mathrm{N'}}[\Eneg[\ell(f(\feature_{\mathrm{N'}},\feature_\mathrm{N}))]] \text{,}\\
\gamma^\ell &= \Epos[\Eneg[\ell(f(\feature_\mathrm{P},\feature_\mathrm{N}))+\ell(f(\feature_\mathrm{N},\feature_\mathrm{P}))]] = B+C\text{,}\\
\gamma^\ell(\feature, \feature') &= \ell(f(\feature, \feature')) + \ell(f(\feature',\feature)) \text{.}
\end{align*}
First, we show that $A =  \E_{\mathrm{P'}}[\Epos[\ell(f(\feature_{\mathrm{P'}},\feature_\mathrm{P}))]] = \E_{\mathrm{P'}}[\Epos[\frac{\gamma^\ell(\feature_{\mathrm{P'}},\feature_\mathrm{P})}{2}]]$:
\begin{align*}
\E_{\mathrm{P'}}[\Epos[\ell(f(\feature_{\mathrm{P'}},\feature_\mathrm{P}))]] &= \E_{\mathrm{P'}}[\Epos[\mathbbm{1}_{\feature_{\mathrm{P'}} = \feature_\mathrm{P}} \ell(0)  + \mathbbm{1}_{\feature_{\mathrm{P'}} \neq \feature_\mathrm{P}} \ell(f(\feature_{\mathrm{P'}},\feature_\mathrm{P})) ]] \\
&=  \E_{\mathrm{P'}}[\Epos[\mathbbm{1}_{\feature_{\mathrm{P'}} = \feature_\mathrm{P}} \ell(0) ]] +  \E_{\mathrm{P'}}[\Epos[ \mathbbm{1}_{\feature_{\mathrm{P'}} \neq \feature_\mathrm{P}} \ell(f(\feature_{\mathrm{P'}},\feature_\mathrm{P})) ]] \\
&= 0 + \E_{\mathrm{P'}}[\Epos[1 \times \ell(f(\feature_{\mathrm{P'}},\feature_\mathrm{P})) ] \\
&= \E_{\mathrm{P'}}[\Epos[\frac{\ell(f(\feature_{\mathrm{P'}},\feature_\mathrm{P}))+  \ell(f(\feature_\mathrm{P}, \feature_{\mathrm{P'}}))}{2}]] \\
&= \E_{\mathrm{P'}}[\Epos[\frac{\gamma^\ell(\feature_{\mathrm{P'}},\feature_\mathrm{P})}{2}]] \text{.}\\
\end{align*}
D can also be rewritten in a same manner so it is omitted for brevity.
\begin{align*}
D =  \E_{\mathrm{N'}}[\Eneg[\ell(f(\feature_{\mathrm{N'}},\feature_\mathrm{N}))]] = \E_{\mathrm{N'}}[\Eneg[\frac{\gamma^\ell(\feature_{\mathrm{N'}},\feature_\mathrm{N})}{2}]]\text{.}
\end{align*}
Then, we get the following result:
\begin{align*}
R^\ell_{\mathrm{AUC}\text{-}\mathrm{Corr}}(g) ={}& \pi \pi' A + (1-\pi)\pi' B + \pi(1-\pi') C + (1-\pi)(1-\pi') D \\ 
={}&\pi \pi' A + (1-\pi)\pi' (\gamma^\ell - C) + \pi(1-\pi') C + (1-\pi)(1-\pi') D \\ 
={}& \pi \pi' A + (\pi'-\pi\pi')\gamma^\ell + (\pi-\pi') C +(1-\pi)(1-\pi')D \\
\begin{split}
 ={}&  (\pi-\pi')R^\ell_{\mathrm{AUC}}(g) + (1-\pi)\pi' \Epos[\Eneg[\gamma^\ell(\feature_\mathrm{P},\feature_\mathrm{N})]] \\ &+ \pi\pi' \E_{\mathrm{P'}}[\Epos[\frac{\gamma^\ell(\feature_{\mathrm{P'}},\feature_\mathrm{P})}{2}]] + (1-\pi)(1-\pi')   \E_{\mathrm{N'}}[\Eneg[\frac{\gamma^\ell(\feature_{\mathrm{N'}},\feature_\mathrm{N})}{2}]] \\ 
\end{split}\\
\begin{split}
 ={}&  (\pi-\pi')R^\ell_{\mathrm{AUC}}(g) + (\pi'-\pi\pi') \Epos[\Eneg[\gamma^\ell(\feature_\mathrm{P},\feature_\mathrm{N})]] \\ &+ \frac{\pi\pi'}{2} \E_{\mathrm{P'}}[\Epos[\gamma^\ell(\feature_{\mathrm{P'}},\feature_\mathrm{P})]] + \frac{(1-\pi)(1-\pi')}{2}   \E_{\mathrm{N'}}[\Eneg[\gamma^\ell(\feature_{\mathrm{N'}},\feature_\mathrm{N})]] \text{.}\\ 
\end{split}
\end{align*}
Therefore, minimizing $R^\ell_{\mathrm{AUC}\text{-}\mathrm{Corr}}(g)$ does not imply minimizing $R^\ell_{\mathrm{AUC}}(g)$ unless $\ell(f(\feature, \feature'))+\ell(f(\feature', \feature))$ is a constant.  
\end{proof}
\subsection{Proof of Theorem 3}
Let $\gamma^\ell(\feature) = \ell(g(\feature)) + \ell(-g(\feature))$, $R^\ell_{\mathrm{BER}\text{-}\mathrm{Corr}}(g)$ can be expressed as

\begin{align*}
R^\ell_{\mathrm{BER}\text{-}\mathrm{Corr}}(g) ={}& (\pi-\pi') {R^\ell_{\mathrm{BER}}(g)}+ \frac{\pi' \Epos[\gamma^\ell(\feature)] + (1-\pi)\Eneg[\gamma^\ell(\feature)]}{2}
\end{align*}

\begin{proof} Recall that the balanced risk is:
\begin{align*}
R^\ell_{\mathrm{BER}}(g) &= \frac{1}{2} \big[\Epos[\ell(g(\feature))\big] + \Eneg\big[\ell(-g(\feature))]\big] \text{.}
\end{align*}
Balanced corrupted risk where $X_\mathrm{CP}$ is assigned to be positive and $X_\mathrm{CN}$ as negative:
\begin{align*}
	R^\ell_{\mathrm{BER}\text{-}\mathrm{Corr}}(g) =\frac{1}{2} \big[ R_{\mathrm{CP}}^{\ell}(g) + R_{\mathrm{CN}}^{\ell}(g) \big] \text{,}
\end{align*}
where
\begin{align*}
R_{\mathrm{CP}}^{\ell}(g) &= \pi \Epos[\ell(g(\feature))]  + (1-\pi)\Eneg[\ell(g(\feature))]  \text{,}\\
R_{\mathrm{CN}}^{\ell}(g) &= \pi' \Epos[\ell(-g(\feature))]  + (1-\pi')\Eneg[\ell(-g(\feature))] \text{.}
\end{align*}
$R^\ell_{\mathrm{BER}\text{-}\mathrm{Corr}}(g)$ can be rewritten as follows:
\begin{align*}
\begin{split}
    2R^\ell_{\mathrm{BER}\text{-}\mathrm{Corr}}(g) ={}& \pi \Epos[\ell(g(\feature))]  + (1-\pi)\Eneg[\ell(g(\feature))] \\
    &+ \pi' \Epos[\ell(-g(\feature))]  + (1-\pi')\Eneg[\ell(-g(\feature))]
\end{split}\\
\begin{split}
    ={}& \pi \Epos[\ell(g(\feature))]  + (1-\pi)\Eneg[\gamma^{\ell}(\feature)-\ell(-g(\feature))] \\ &+ \pi' \Epos[\gamma^{\ell}(\feature)-\ell(g(\feature))]  + (1-\pi')\Eneg[\ell(-g(\feature))]
\end{split}\\
\begin{split}
    ={}& \pi \Epos[\ell(g(\feature))] + (1-\pi)\Eneg[\gamma^{\ell}(\feature)] - (1-\pi)\Eneg[\ell(-g(\feature))]\\ &+ \pi'\Epos[\gamma^{\ell}(\feature)] - \pi' \Epos[\ell(g(\feature))] + (1-\pi')\Eneg[\ell(-g(\feature))]
\end{split}\\
\begin{split}
    ={}& \pi \Epos[\ell(g(\feature))] - \pi' \Eneg[\ell(-g(\feature))] + \pi \Eneg[\ell(-g(\feature))] \\ &- \pi' \Epos[\ell(g(\feature))] + (1-\pi) \Eneg[\gamma^{\ell}(\feature)] + \pi' \Epos[\gamma^{\ell}(\feature)]
\end{split}\\
    ={}& (\pi-\pi') [ \Epos[\ell(g(\feature))] + \Eneg[\ell(-g(\feature))]] + \pi' \Epos[\gamma^{\ell}(\feature)] + (1-\pi)\Eneg[\gamma^{\ell}(\feature)]\\
={}& 2(\pi-\pi') R^\ell_{\mathrm{BER}}(g) + \pi' \Epos[\gamma^{\ell}(\feature)] + (1-\pi)\Eneg[\gamma^{\ell}(\feature)] \\
R^\ell_{\mathrm{BER}\text{-}\mathrm{Corr}}(g) ={}& (\pi-\pi') {R^\ell_{\mathrm{BER}}(g)}+ \frac{\pi' \Epos[\gamma^\ell(\feature)] + (1-\pi)\Eneg[\gamma^\ell(\feature)]}{2} \text{.}
\end{align*}
\end{proof}
\subsection{Conditional risk for binary classification}
By making use of the symmetric property, i.e., $\ell(z)+\ell(-z)= K$, a pointwise conditional risk can be rewritten such that there is only one term depending on $\alpha$ as follows for a fixed $\feature$:
\begin{align*}
	C^\ell_\eta(\alpha) &= \eta \ell(\alpha) + (1-\eta)\ell(-\alpha) \\
    &= \eta \ell(\alpha) + (1-\eta)(K-\ell(\alpha)) \\
    &= (1-\eta)K + (2\eta-1)\ell(\alpha)\text{,}
\end{align*}
where $\eta = p(y=1|\feature)$. It can be observed that $\ell(-z)$ can be expressed by $K-\ell(z)$.
The symmetric property makes analysis simpler because $\ell(-z)$ can be rewritten as $\ell(z)$and the following general properties can be obtained by only rely on the symmetric property.
\subsection{Proof of Theorem 5}
\begin{proof}

Let $H(\eta) = \inf\limits_{\alpha \in \mathbb{R}} C^\ell_\eta(\alpha)$ and $H^{-}(\eta) = \inf\limits_{\alpha:\alpha(2\eta-1)\leq 0} C^\ell_\eta(\alpha)$ \text{.}

First, consider the $\psi$-transform from the definition 2 of~\citet{bartlett2006}. Consider $\ell\colon\R \to [0,\infty)$, function $\psi:[0,1] \to [0,\infty)$ by $\psi=\tilde{\psi}^{**}$, where

\begin{align*}
\tilde{\psi}(\theta) = H^{-}\bigg(\frac{1+\theta}{2}\bigg) - H\bigg(\frac{1+\theta}{2}\bigg),
\end{align*}
$g^{**}\colon [0,1] \to \R$ is the Fenchel-Legendre biconjugate of $g:[0,1] \to \R$ characterized by
\begin{align*}
\mathrm{epi} \, g^{**} =\overline{co} \, \mathrm{epi} \, g.
\end{align*}
It is known that $\psi=\tilde{\psi}$ if and only if $\tilde{\psi}$ is convex. For more details, please refer to~\citet{bartlett2006}.

Next, we use the following statements in Lemma 5 from~\citet{bartlett2006} which can be interpreted that,
$\ell$ is classification-calibrated if and only if $\psi(\theta) > 0$ for all $ \theta \in (0,1]$. Based on this statement, we prove the sufficient and necessary condition for symmetric losses to be classification-calibrated by showing that $\psi(\theta) > 0$ for all $ \theta \in (0,1]$ if and only if $\inf\limits_{\alpha > 0} \ell(\alpha) < \inf\limits_{\alpha \leq 0}  \ell(\alpha)$.

Using the conditional risk of symmetric losses in the previous section, $H$ and $H^{-}$ can be written as 
\begin{align*}
H(\eta) &= \inf\limits_{\alpha \in \mathbb{R}}  C^\ell_\eta(\alpha) \\
&= (1-\eta)K + \inf\limits_{\alpha \in \mathbb{R}}(2\eta -1) \ell(\alpha) \text{,}\\
H^{-}(\eta) &=  \inf\limits_{\alpha:\alpha(2\eta-1)\leq 0} C^\ell_\eta(\alpha) \\
&= (1-\eta)K + \inf\limits_{\alpha:\alpha(2\eta-1)\leq 0} (2\eta -1) \ell(\alpha)\text{.}
\end{align*}
Let $\tilde{\psi}(\theta) = H^{-}(\frac{1+\theta}{2}) -  H(\frac{1+\theta}{2})$ where $ \theta \in (0,1]$\text{,}
\begin{align*}
\tilde{\psi}(\theta) &= H^{-}(\frac{1+\theta}{2}) -  H(\frac{1+\theta}{2}) \\
&= \inf\limits_{\alpha:\alpha \theta \leq 0} \theta \ell(\alpha) - \inf\limits_{\alpha \in \mathbb{R}}  \theta  \ell(\alpha)\\
& = \theta [\inf\limits_{\alpha \leq 0}  \ell(\alpha) - \inf\limits_{\alpha \in \mathbb{R}}   \ell(\alpha)]\text{.}
\end{align*}
Let $  \mathrm{C} = \inf\limits_{\alpha \leq 0}  \ell(\alpha) - \inf\limits_{\alpha \in \mathbb{R}}   \ell(\alpha)$ is a constant depends on the function. 
\begin{align*}
\tilde{\psi}(\theta) &= C\theta\text{.}
\end{align*}
Here, $\tilde{\psi}(\theta)$ is linear and therefore convex. As a result, $\psi = \tilde{\psi}$.
Based on Lemma 5 of~\citet{bartlett2006}. $\ell$ is classification-callibrated if and only if $\psi(\theta) > 0$ for all $\theta \in (0,1].$ In this case, $\theta$ is positive therefore, any symmetric loss function is classification-calibrated if and only if $C > 0$.
\begin{align*}
\inf\limits_{\alpha \leq 0}  \ell(\alpha) - \inf\limits_{\alpha \in \mathbb{R}}   \ell(\alpha) &> 0 \\
 \inf\limits_{\alpha \in \mathbb{R}} \ell(\alpha) &< \inf\limits_{\alpha \leq 0}  \ell(\alpha) \\
  \inf\limits_{\alpha > 0} \ell(\alpha) &< \inf\limits_{\alpha \leq 0}  \ell(\alpha)\text{.}
\end{align*}
Therefore, a symmetric loss $\ell$ is classification-calibrated if and only if $ \inf\limits_{\alpha > 0} \ell(\alpha) < \inf\limits_{\alpha \leq 0}  \ell(\alpha)$.
\end{proof}
\subsection{Proof of Theorem 7}
\begin{proof}
Once $\psi(\theta) =  [\inf\limits_{\alpha \leq 0}  \ell(\alpha) - \inf\limits_{\alpha > 0} \ell(\alpha)]\theta$ is obtained in the previous proof of classification-calibration for a symmetric loss. It is straightforward to obtain an excess risk bound based on~\citet{bartlett2006}:
\begin{align*}
	\psi(R^{\zerooneloss}(g)-R^{\zerooneloss*}) &\leq R^{\ell}(g) - R^{\ell*} \\
	[ \inf\limits_{\alpha \leq 0}  \ell(\alpha) - \inf\limits_{\alpha > 0} \ell(\alpha)](R^{\zerooneloss}(g)-R^{\zerooneloss*}) &\leq R^{\ell}(g) - R^{\ell*} \\
	R^{\zerooneloss}(g)-R^{\zerooneloss*} &\leq \frac{ R^{\ell}(g) - R^{\ell*}}{\inf\limits_{\alpha \leq 0}  \ell(\alpha) - \inf\limits_{\alpha > 0} \ell(\alpha)} \text{,}
\end{align*}
where $R^{\ell*} = \inf\limits_{g} R^{\ell}(g)$ and $R^{\zerooneloss*} = \inf\limits_{g} R^{\zerooneloss}(g)$.
\end{proof}
\subsection{Proof of Theorem 8}
\begin{proof}
Consider a conditional risk minimizer of a symmetric loss $\ell$
\begin{align*}
f_{\ell}^*(\feature) &= \argmin\limits_{\alpha \in \R} C^{\ell}_{\eta(\feature)}(\alpha) \\
&= \argmin\limits_{\alpha \in \R}  (1-\eta(\feature))K + (2\eta(\feature)-1)\ell(\alpha) \text{.}
\end{align*}
The constants can be ignored as it does not depend on $\alpha$. Let us consider two cases of $\eta > \frac{1}{2}$ and $\eta < \frac{1}{2}$:

\noindent Case 1: $\eta > \frac{1}{2}$
\begin{align*}
f_{\ell}^*(\feature) &= (1-\eta(\feature))K + \argmin\limits_{\alpha \in \R}  (2\eta(\feature)-1)\ell(\alpha) \\
&= \argmin\limits_{\alpha \in \R} \ell(\alpha) \text{.}
\end{align*}
Case 2: $\eta < \frac{1}{2}$
\begin{align*}
f_{\ell}^*(\feature) &=  (1-\eta(\feature))K +  \argmin\limits_{\alpha \in \R}  (2\eta(\feature)-1)\ell(\alpha) \\
&= \argmax\limits_{\alpha \in \R} \ell(\alpha) \text{.}
\end{align*}
Suppose there are many $\alpha$ to satisfy the conditions. Due to the symmetric condition, We can express the following relations. 
\begin{align*}
	\argmin\limits_{\alpha \in \R} \ell(\alpha) = - \argmax\limits_{\alpha \in \R} \ell(\alpha)\text{,}
\end{align*}
where $-\argmax\limits_{\alpha \in \R} \ell(\alpha)$ means a set such that each element in the set $\argmax\limits_{\alpha \in \R} \ell(\alpha)$ is multiplied by $-1$. As a result, $f_{\ell}^*(\feature)$ can be simply written as follows:
\begin{align*}
f_{\ell}^*(\feature) = \mathrm{M \, sign}(\eta(\feature)-\frac{1}{2}) \text{,}
\end{align*}
where M  $\in \argmin\limits_{\alpha \in \R}\ell(\alpha)$. This result shows that the conditional risk minimizer of a symmetric loss can be expressed as the bayes classifier scaled by a constant. In the case of functions such that it is classification-calibrated and argmin cannot be obtained, $M \to \infty$.
\end{proof}
\subsection{Introduction of AUC-consistency}
In AUC maximization, we want to find the function $g$ that minimizes the following risk:
\begin{align*}
R^{\zerooneloss}_{\mathrm{AUC}}(g) = \Epos [ \Eneg[\zerooneloss(g(\feature_\mathrm{P}) - g(\feature_\mathrm{N}))]] \text{.}
\end{align*}
\citet{gao2015consistency} showed that the Bayes optimal functions can be expressed as follows:
\begin{align*}
	\mathcal{B} &= \{ g: R^{\zerooneloss}_{\mathrm{AUC}}(g) = R^{\zerooneloss*}_{\mathrm{AUC}}(g) \} \\
	&= \{ g:(g(\feature)-g(\feature')) (\eta(\feature) - \eta(\feature')) > 0 \, \mathrm{if} \, \eta(\feature) \neq \eta(\feature') \}
\end{align*}
Unlike classification-calibration, the Bayes optimal functions for AUC maximization depend on the \emph{pairwise} class probability, i.e., the class probabilities for two data points are compared. The optimal function $g$ is a function such that the sign of $g(\feature)-g(\feature')$ matches the sign of $\eta(\feature) - \eta(\feature')$. Therefore, one solution of $g$ is the class probability itself. Because when $g(\feature) = \eta(\feature)$ for all $\feature$, then $g(\feature)-g(\feature') = \eta(\feature) - \eta(\feature')$ which is exactly the same value as the function we want to match the sign with. As a result, it is arguable that the bipartite ranking problem based on the AUC score is easier than the class conditional probability estimation problem in the sense that the problem is solved if we have an access to $\eta(\feature)$. However, we only need to find a function $g$ such that $\mathrm{sign}(g(\feature)-g(\feature')) = \mathrm{sign}(\eta(\feature) - \eta(\feature'))$. 
\emph{AUC-consistency} property can be treated as the minimum requirement of a loss function to be suitable for bipartite ranking~\citep{gao2015consistency}.

\subsection{Proof of Lemma 9}
A proof is based on a necessary of the notion of calibration in~\citet{gao2015consistency}, which we call AUC-calibration to avoid confusion in this paper.
According to~\citet{gao2015consistency}, AUC-calibration is a necessary condition for AUC-consistency.
Here, we prove that a symmetric loss is AUC-calibrated if and only if a symmetric loss is classification-calibrated.
\begin{proof}
For a symmetric loss $\ell$, we can rewrite a pairwise conditional risk term in the infimum as follows:
\begin{align*}
 \eta(1-\eta')\ell(\alpha) + \eta'(1-\eta)\ell(-\alpha) &=   \eta(1-\eta')\ell(\alpha) + \eta'(1-\eta)(K - \ell(\alpha))\\
 &=  \eta(1-\eta')\ell(\alpha) + \eta'K(1-\eta) - \eta' \ell(\alpha) + \eta\eta' \ell(\alpha) \\
 &= (\eta-\eta')\ell(\alpha) + \eta'K(1-\eta)\text{.}
\end{align*}

\begin{align*}
	H^{-}(\eta, \eta') &> H(\eta, \eta') \\
	H^{-}(\eta, \eta') - H(\eta, \eta') &> 0 \\
	\frac{1}{2\pi(1-\pi)} [ \inf\limits_{\alpha:\alpha(\eta-\eta')\leq 0} (\eta-\eta') \ell(\alpha) -  \inf\limits_{\alpha \in \R} (\eta-\eta') \ell(\alpha)] &> 0 \\
	 \inf\limits_{\alpha:\alpha(\eta-\eta')\leq 0} (\eta-\eta') \ell(\alpha) -  \inf\limits_{\alpha \in \R} (\eta-\eta') \ell(\alpha) &> 0
\end{align*}
Case 1: $\eta - \eta' > 0$
\begin{align*}
		 (\eta-\eta')  [ \inf\limits_{\alpha:\alpha(\eta-\eta')\leq 0}\ell(\alpha) -  \inf\limits_{\alpha \in \R} \ell(\alpha)] &> 0 \\
		   \inf\limits_{\alpha:\alpha\leq 0}\ell(\alpha) -  \inf\limits_{\alpha \in \R} \ell(\alpha) > 0\text{.}
\end{align*}
Case 2: $\eta - \eta' < 0$
\begin{align*}
		 (\eta-\eta')  [ \sup\limits_{\alpha:\alpha(\eta-\eta')\leq 0}\ell(\alpha) -  \sup\limits_{\alpha \in \R} \ell(\alpha)] > 0 \\
		  \sup\limits_{\alpha:\alpha(\eta-\eta')\leq 0}\ell(\alpha) -  \sup\limits_{\alpha \in \R} \ell(\alpha) < 0 \\
		   \sup\limits_{\alpha:\alpha \geq 0}\ell(\alpha) -  \sup\limits_{\alpha \in \R} \ell(\alpha) < 0\text{.}
\end{align*}
The two inequalities are equivalent which proved in Section 4.9.6. Therefore, a symmetric loss must satisfy $\inf\limits_{\alpha> 0} \ell(\alpha)< \inf\limits_{\alpha\leq 0} \ell(\alpha)$
to be AUC-calibrated. 
This is equivalent to classification-calibration condition for a symmetric loss.
Next, it is known that AUC-calibration is a necessary condition for AUC-consistent~\citep{gao2015consistency}, therefore, a symmetric loss that is not classification-calibrated must not satisfy this condition, and thus not AUC-consistent.

This elucidates that classification-calibration is a necessary condition for a symmetric loss to be AUC-consistent. 
\end{proof}
\subsection{Proof of Proposition 10}
\begin{proof}
Consider a pairwise conditional risk:
\begin{align}\label{eq:auc-cond-risk}
 \eta(1-\eta')\ell(\alpha) + \eta'(1-\eta)\ell(-\alpha) &=   \eta(1-\eta')\ell(\alpha) + \eta'(1-\eta)(K - \ell(\alpha)) \nonumber\\
 &=  \eta(1-\eta')\ell(\alpha) + \eta'K(1-\eta) - \eta' \ell(\alpha) + \eta\eta' \ell(\alpha) \nonumber\\
 &= (\eta-\eta')\ell(\alpha) + \eta'K(1-\eta)\text{.}
\end{align}

 Then, let us consider a symmetric loss $\ell_{\mathrm{EX}}$ such that $\ell_{\mathrm{EX}}(1)=~0$, $\ell_{\mathrm{EX}}(-1)=1,$ and $0.5$ otherwise. 
 It is straightforward to see that it is a symmetric loss where $\ell_{\mathrm{EX}}(z) + \ell_{\mathrm{EX}}(-z) = 1$. We are going to show that this loss is classification-calibrated but AUC-consistent. 
 Moreover, we can see that $ \inf\limits_{\alpha > 0} \ell_{\mathrm{EX}}(\alpha) < \inf\limits_{\alpha \leq 0}  \ell_{\mathrm{EX}}(\alpha)$. Therefore, $\ell_{\mathrm{EX}}$ is classification-calibrated based on the previous theorem on a necessary and sufficient condition of a symmetric loss to be classification-calibrated.
 
 Next, let us consider a uniform discrete distribution $D_\mathrm{U}$ that contains 3 possible supports $\{\feature_1, \feature_2, \feature_3\}$.
 Moreover, let $\eta(\feature_1) = 1$, $\eta(\feature_2)=0.5$, $\eta(\feature_3)=0$.
 
 Here, we prove Proposition 10 by a counterexample that the minimizer of the AUC risk with respect to $\ell_{\mathrm{EX}}$ resulted in a function that behaves differently the Bayes-optimal solution of AUC maximization of a function that has a strictly monotonic relationship with the class probability $\eta(\feature)$~\citep{menon2016bipartite}, and therefore AUC-inconsistent.
 
 Consider the following pairwise risk:
\begin{align*}
R^{\mathrm{pair}}_{\ell_{\mathrm{EX}}}(g) &= \frac{1}{2\pi(1-\pi)} \mathop{\E}\limits_{\feature,\feature' \sim D^2_X}[\eta(\feature)(1-\eta(\feature')) \ell_{\mathrm{EX}}(g(\feature)-g(\feature')) \\ &\quad + \eta(\feature')(1-\eta(\feature)\ell_{\mathrm{EX}}(g(\feature')-g(\feature))]\\
&= \frac{1}{2\pi(1-\pi)} \mathop{\E}\limits_{\feature,\feature' \sim D^2_X}[(\eta(\feature)-\eta(\feature'))\ell_{\mathrm{EX}}(g(\feature)-g(\feature')) + \eta'K(1-\eta)]
\end{align*}

Since we are only interested in the minimizer of the risk, let us ignore the constant term and rewrite the risk pair as follows:
\begin{align*}
R^{\mathrm{pair}}_{\ell_{\mathrm{EX}}}(g) &= C_0 + C_1 \sum_{i=1}^{3} \sum_{j \neq i} \ell_{\mathrm{EX}}(g(\feature_i)-g(\feature_j)),
\end{align*}
where $C_0$ and $C_1$ are some constants.

Let us consider the following $g_1$, $g_2$, $g_3$, $g_4$: $\R^d \to \R$,
\begin{align*}
    g_1(\feature_1)&=g_1(\feature_2)+1=g_1(\feature_3)+1 , \\
    g_2(\feature_1)&=g_2(\feature_2)+1=g_2(\feature_3)+2 ,\\
    g_3(\feature_1)&=g_3(\feature_2)=g_3(\feature_3) ,\\
    g_4(\feature_1)&=g_4(\feature_2)=g_4(\feature_3)+1.\\
\end{align*}

Then, a function $g$ that minimizes the risk $R^{\mathrm{pair}}_{\ell_{\mathrm{EX}}}(g)$ is the one that minimizes $\sum_{i=1}^{3} \sum_{j \neq i} \ell_{\mathrm{EX}}(g(\feature_i)-g(\feature_j)) = \sum_{i=1}^{3} \sum_{j \neq i} \ell_{\mathrm{EX}}(g(\feature_i,\feature_j))$.
More precisely, there are six pairs to consider as can be observed in the following table.

\begin{table}[ht]
\centering
\caption{The illustrations of the values for each pair in the uniform discrete distribution supports.} 
\label{table:counter-example}
\vspace{0.05in}
\begin{tabular}{|C|C|C | C | C | C | C|C|}
\hline
\text{Pair} & \eta_i - \eta_j &  \ell_{\mathrm{EX}}(g_1(\feature_i,\feature_j)) & \ell_{\mathrm{EX}}(g_2(\feature_i,\feature_j)) & \ell_{\mathrm{EX}}(g_3(\feature_i,\feature_j)) & \ell_{\mathrm{EX}}(g_4(\feature_i,\feature_j))\\ \hline
\eta_1 - \eta_2 & 0.5&0&  0 &0.5&  0.5\\ 
\eta_1 - \eta_3 & 1 &0&  0.5&0.5&0\\ 
\eta_2 - \eta_1 & -0.5 &1&  1 &0.5 &  0.5\\ 
\eta_2 - \eta_3 & 0.5 &0.5&  0 &0.5 &0\\ 
\eta_3 - \eta_1  & -1 &1&0.5 &  0.5 &1\\ 
\eta_3 - \eta_2& -0.5 &0.5& 1 &  0.5 &  1 \\ 
     \hline
\end{tabular}
\end{table}

We can rank the score of each $g$ by taking a weighted sum of column "$\eta_i - \eta_j$" in Table~\ref{table:counter-example} to the column of the loss function of a function $g$. For example, for $g_1$, the score is $0.5 \times 0 + 1 \times 0 + (-0.5)\times1 + 0.5 \times 0.5 + (-1) \times 1 + (-0.5) \times 0.5 = -1.5$. Note that the lower sum the better since we are interested in the minimizer.

The function $g_2$ is a function that is optimal with respect to the pairwise risk with respect to the zero-one loss, i.e., has a strictly monotonic relationship with the class probability $\eta(\feature)$. However, the score of $g_2$ is $-1$ which is worse than $g_1$ and $g_4$. In this scenario, $g_1$ and $g_4$ minimize the risk in this distribution which contradicts to the optimal solution of AUC optimization.

Note that $g_1$ and $g_4$ are the global minimizer of the risk, not only among $g_1, g_2, g_3, g_4$. 
Since $\ell_{\mathrm{EX}}$ returns the same value of all input \emph{except two points} which are $1$ and $-1$, the minimizer of the risk is the one that the loss function returns $1$ for the lowest weight, i.e., for $\eta_i - \eta_j = -1$ and $\eta_i - \eta_j = -0.5$. 

Intuitively, to fill in the blanks for all pairs, once we pick where the loss will return $1$ for two pairs, all other pairs will be fixed. For other terms, they will cancel each other out and therefore the variable term minimum pairwise risk in the distribution $D$ with respect to the loss $\ell_{\mathrm{EX}}$ is $-1.5$, which includes the one that is not the Bayes-optimal solution and the one that conforms to the Bayes-optimal solution is not included. 

Thus, we conclude that $\ell_{\mathrm{EX}}$, which is a classification-calibrated symmetric loss is AUC-inconsistent. This suggests the gap between classification calibration and AUC-consistency for a symmetric loss.
\end{proof}

\subsection{Proof of Theorem 11}
\begin{proof}
Recall the Bayes optimal functions for AUC-optimization~\citet{gao2015consistency} :
\begin{align*}
	\mathcal{B} &= \{ g: R^{\zerooneloss}_{\mathrm{AUC}}(g) = R^{\zerooneloss*}_{\mathrm{AUC}}(g) \} \\
	&= \{ g:(g(\feature)-g(\feature')) (\eta(\feature) - \eta(\feature')) > 0 \, \mathrm{if} \, \eta(\feature) \neq \eta(\feature') \}.
\end{align*}

Here, we consider $\ell$ as a non-increasing loss $\ell\colon \R \to \R$ such that $\ell(z)+\ell(-z)$ is a constant and $\ell'(0) < 0$.

Let us write
\begin{align*}
R^{\mathrm{pair}}_{\ell}(g) &= \frac{1}{2\pi(1-\pi)} \mathop{\E}\limits_{\feature,\feature' \sim D^2_X}[\eta(\feature)(1-\eta(\feature')) \ell(g(\feature)-g(\feature')) \\ &\quad + \eta(\feature')(1-\eta(\feature))\ell(g(\feature')-g(\feature))]\\
&= \frac{1}{2\pi(1-\pi)} \mathop{\E}\limits_{\feature,\feature' \sim D^2_X}[(\eta(\feature)-\eta(\feature'))\ell(g(\feature)-g(\feature')) + \eta'K(1-\eta)].
\end{align*}

Next, we show that the minimizer of the AUC risk of $\ell$, has a strictly monotonic relationship with the class probability $\eta(\feature)$.
More precisely, we will prove the following inequality:
\begin{align}\label{proof:sym-auccon}
    \inf\limits_{g \notin \mathcal{B}}R^{\mathrm{pair}}_{\ell}(g) > \inf\limits_{g} R^{\mathrm{pair}}_{\ell}(g) .
\end{align}

We will prove by contradiction. First, let us assume that there is a function $g_\mathrm{B}$ that is not strictly monotonic to the class probability $\eta(\feature)$ but is a minimizer of the AUC risk $R^{\mathrm{pair}}_{\ell}$.
Then, we prove that it is impossible since there always exists a function that can further minimize the AUC risk AUC risk $R^{\mathrm{pair}}_{\ell}$.
Note that the key idea of the proof is similar to that of \citet{gao2015consistency} except the fact that a loss is not convex and we can make use of the symmetric property.

First, similarly to the proof of the previous proposition, by making use of symmetric property, let $C_0,C_1,C_2,C_3$ be some constants, we obtain the following
\begin{align} \label{eq:sym-auc-risk}
R^{\mathrm{pair}}_{\ell}(g) &= \frac{1}{2\pi(1-\pi)} \mathop{\E}\limits_{\feature,\feature' \sim D^2_X}[(\eta(\feature)-\eta(\feature'))\ell(g(\feature)-g(\feature')) + \eta'K(1-\eta)]. \nonumber\\
&= C_0 \mathop{\E}\limits_{\feature,\feature' \sim D^2_X}[(\eta(\feature)-\eta(\feature'))\ell(g(\feature)-g(\feature'))] + C_1. \nonumber\\
&= C_0 \mathop{\E}\limits_{\feature,\feature' \sim D^2_X, \eta(\feature)>\eta(\feature')}\left[(\eta(\feature)-\eta(\feature'))\left(\ell(g(\feature)-g(\feature')) - \ell(g(\feature)-g(\feature')) \right)\right] + C_1. \nonumber\\
&= C_0 \mathop{\E}\limits_{\feature,\feature' \sim D^2_X, \eta(\feature)>\eta(\feature')}\left[(\eta(\feature)-\eta(\feature'))\left(2\ell(g(\feature)-g(\feature')) - K \right)\right] + C_1. \nonumber\\
&= C_2 \mathop{\E}\limits_{\feature,\feature' \sim D^2_X, \eta(\feature)>\eta(\feature')}\left[(\eta(\feature)-\eta(\feature'))\ell(g(\feature)-g(\feature'))\right] + C_3.
\end{align}
The key advantage for the symmetric loss is that there is only one term that involves a loss for each pair $\ell(g(\feature)-g(\feature'))$, this helps us handle the conditional risk easier similarly to the binary classification scenario.

Next, we will show that for any $g_\mathrm{B} \notin \mathcal{B} $ there exists a better function $g_{\mathrm{G}}$ such that
\begin{align}
\label{eq:risk-target-thm-11}
    R^{\mathrm{pair}}_{\ell}(g_\mathrm{B}) > R^{\mathrm{pair}}_{\ell}(g_\mathrm{G}).
\end{align}

By ignoring constants, the term that a function $g$ can minimize the risk for a symmetric loss is 
\begin{align*}
    R^{\mathrm{comp}}_{\ell}(g) = \mathop{\E}\limits_{\feature,\feature' \sim D^2_X, \eta(\feature)>\eta(\feature')}\left[(\eta(\feature)-\eta(\feature'))\ell(g(\feature)-g(\feature'))\right]
\end{align*}

To show that \eqref{eq:risk-target-thm-11} holds, it suffices to show that
\begin{align}
\label{eq:comp-analysis}
    R^{\mathrm{comp}}_{\ell}(g_\mathrm{B}) > R^{\mathrm{comp}}_{\ell}(g_\mathrm{G})
\end{align}

Then, we know that there exists $\feature_1$ and $\feature_2$, which is a pair such that $g_\mathrm{B}(\feature_1) \leq g_\mathrm{B}(\feature_2)$, but $\eta(\feature_1) > \eta(\feature_2)$. Let $\delta = |g_\mathrm{B}(\feature_1)-g_\mathrm{B}(\feature_2) |+ \epsilon$, where $\epsilon > 0$.

Let us construct $g_\mathrm{G}$ as follows.
\begin{align*}
    g_\mathrm{G}(\feature) &= g_\mathrm{B}(\feature) - \delta\text{, if } \eta(\feature) \leq \eta(\feature_1) \\
    g_\mathrm{G}(\feature) &= g_\mathrm{B}(\feature) + \delta\text{, if } \eta(\feature) > \eta(\feature_1)
\end{align*}

Since $\eta(\feature_1)>\eta(\feature_2)$, $g_\mathrm{B}(\feature_1)-g_\mathrm{B}(\feature_2) \leq 0$, $g_\mathrm{G}(\feature_1)-g_\mathrm{G}(\feature_2) > 0$, and $\ell$ is non-increasing and $\ell'(0)<0$, it is straightforward to see that
\begin{align}
\label{eq:good-function-win}
    (\eta(\feature_1)-\eta(\feature_2))\ell(g_\mathrm{B}(\feature_1)-g_\mathrm{B}(\feature_2)) > (\eta(\feature_1)-\eta(\feature_2))\ell(g_\mathrm{G}(\feature_1)-g_\mathrm{G}(\feature_2)) .
\end{align}

Next, we show that modifications of other pairs from the construction of $g_\mathrm{G}$ will not further increase the $R^{\mathrm{comp}}_{\ell}$ with respect to $R^{\mathrm{comp}}_{\ell}(g_\mathrm{B})$. There are three following cases to consider.

Case 1: $A_1$ =$\{\feature \text{ such that } \eta(\feature) > \eta(\feature_1)\}$. Since all $\feature \in A_1$ are modified equally, i.e.,  $g_\mathrm{G}(\feature) = g_\mathrm{B}(\feature) + \delta$. For all $\feature, \feature' \in A_1$
\begin{align*}
(\eta(\feature)-\eta(\feature'))\ell(g_\mathrm{B}(\feature)-g_\mathrm{B}(\feature'))
&=(\eta(\feature)-\eta(\feature'))\ell((g_\mathrm{B}(\feature)+\delta)-(g_\mathrm{B}(\feature')+\delta)) \\
&= (\eta(\feature)-\eta(\feature'))\ell(g_\mathrm{G}(\feature)-g_\mathrm{G}(\feature'))
\end{align*}

Case 2: $A_2$ = $\{\feature \text{ such that } \eta(\feature) \leq \eta(\feature_1)\}$. Since all $\feature \in A_2$ are modified equally, i.e.,  $g_\mathrm{G}(\feature) = g_\mathrm{B}(\feature) - \delta$. For all $\feature, \feature' \in A_2$
\begin{align*}
(\eta(\feature)-\eta(\feature'))\ell(g_\mathrm{B}(\feature)-g_\mathrm{B}(\feature'))
&=(\eta(\feature)-\eta(\feature'))\ell((g_\mathrm{B}(\feature)-\delta)-(g_\mathrm{B}(\feature')-\delta)) \\
&= (\eta(\feature)-\eta(\feature'))\ell(g_\mathrm{G}(\feature)-g_\mathrm{G}(\feature') )
\end{align*}

Case 3: For all $\feature \in A_1$ and $\feature' \in A_2$. Since $\ell$ is a non-increasing function and $\delta > 0$.
\begin{align*}
(\eta(\feature)-\eta(\feature'))\ell(g_\mathrm{B}(\feature)-g_\mathrm{B}(\feature'))
&\geq (\eta(\feature)-\eta(\feature'))\ell(g_\mathrm{B}(\feature)+\delta-g_\mathrm{B}(\feature')+\delta) \\
&=  (\eta(\feature)-\eta(\feature'))\ell(g_\mathrm{B}(\feature)-g_\mathrm{B}(\feature') +2\delta)
\end{align*}

Therefore, with the strict inequality $\eqref{eq:good-function-win}$ and other pairs will not further increase the risk higher than a bad function as shown in the analysis of three cases, we show that \eqref{eq:comp-analysis} must hold, and therefore \eqref{eq:risk-target-thm-11} and \eqref{proof:sym-auccon} hold. As a result, it is impossible that  $\inf\limits_{g \notin \mathcal{B}}R^{\mathrm{pair}}_{\ell}(g) = \inf\limits_{g} R^{\mathrm{pair}}_{\ell}(g)$ since we can always find a better function $g_\mathrm{G}$ compared with a function $g_\mathrm{B} \notin \mathcal{B}$.

Thus, we conclude that \eqref{proof:sym-auccon} holds. Once we show that \eqref{proof:sym-auccon} holds, we can directly use the results from the proof of Theorem 2 in \citet{gao2015consistency} without modification to show that $\ell$ is AUC-consistent.
\end{proof}

Note that we can further relax the condition $\ell'(0) < 0$, we only have to make sure a loss is not a constant function. 
Nevertheless, we prove this condition for $\ell'(0) < 0$ since this is not difficult to satisfy in practice and covers many surrogate losses in the literature to the best of our knowledge.

\section{Details of Implementation and Datasets}
\subsection{Experiments on UCI and LIBSVM Datasets}
We used nine datasets, namely \emph{spambase}, \emph{phoneme}, \emph{phishing}, \emph{phishing}, \emph{waveform}, \emph{susy}, \emph{w8a}, \emph{adult}, \emph{twonorm}, \emph{mushroom}.
We used the one hiddent layer multilayer perceptron as a model ($d-500-1$). 
We used 500 corrupted positive data, 500 corrupted negative data, and balanced 500 test data. 
The corruption for the training data can be done manually by simply mixing positive and negative data according to the class prior of the corrupted positive and corrupted negative data, i.e., $\pi$ and $\pi'$. 
We used rectifier linear units (ReLU) \cite{nair2010rectified}. Learning rate was set to $0.001$, batch size was $500$, and the number of epoch was $100$. 
We ran 20 trials for each experiment and reported the mean values and standard error. 
The objective functions of the neural networks were optimized using AMSGRAD~\citep{adam_amsgrad}. The experiment code was implemented with Chainer~\citep{tokui2015chainer}.

\subsection{Experiments on MNIST and CIFAR-10}
\textbf{MNIST:} The MNIST dataset contains 60,000 gray-scale training images and 10,000 test images from digits 0 to 9. In this experiment which consider the binary classification, we used even and odd digits as positive and negative classes respectively. To make sure same data were not used as both positive and negative class, we sampled 15,000 images for each class. For instance, when noise rate is $(\pi=0.7,\pi'=0.4)$, positive class consists of 10,500 even digits images and 4,500 odd digits images and negative class consists of 6,000 even digits images and 9,500 odd digits images respectively. The model used for MNIST was convolutional neural networks which is same architecture of \citet{ishida2}: d-Conv[18,5,1,0]-Max[2,2]-Conv[48,5,1,0]-Max[2,2]-800-400-1, where Conv[18, 5, 1, 0] means 18 channels of 5$\times$5 convolutions with stride 1 and padding 0, and Max[2,2] means max pooling with kernel size 2 and stride 2. We used rectifier linear units (ReLU) \cite{nair2010rectified} as activation function after fully connected layer followed by dropout layer \cite{srivastava2014dropout} in the first two fully connected layer.

\textbf{CIFAR-10:} The CIFAR-10 dataset contains natural RGB images from 10 classes with 5,000 training images and 1,000 test images per class. Following \citet{ishida2}, we set a class 'airplane' as the positive class and set one of other classes as negative class in order to construct binary classification problem. 
Thus, we conducted experiments on 9 pairs of airplane vs others. 
To make sure same data were not used as both positive and negative class, we sampled 4,540 images for each class. Note that we have a few data differently from MNIST, 4,540 is the highest number we can sure that same data were not duplicated. Same architecture of CNNs was used for experiment of CIFAR-10.

\section{Additional Experimental Results}
In this section, we show the experimental results on additional datasets from the main body.


\subsection{BER Optimization Using UCI and LIBSVM Datasets}
Outperforming methods are highlighted in boldface using one-sided t-test with the significance level 5\%. The experiments were conducted 20 times.
\begin{table}[htb]
\centering
\caption{Mean balanced accuracy and standard error for BER minimization from corrupted labels, where $\pi = 1.0$ and $\pi'=0$.}
\begin{tabular} { |L|L|L|L|L|L|L|L|L|L|L|L|L|L|L|L|L|L|L|L|L|L|}
\hline
\text{Dataset} & \text{Dim.}& \text{Barrier}& \text{Unhinged}& \text{Sigmoid}& \text{Logistic}& \text{Hinge}& \text{Squared}& \text{Savage}\\ \hline
\text{spambase} & 57 &89.4 (0.3) &89.0 (0.3) &90.9 (0.2) &92.2 (0.2) & \textbf{92.2 (0.3)} & \textbf{92.9 (0.3)} & \textbf{92.5 (0.2)}\\ 
\text{phoneme} & 5 &75.2 (0.4) &76.4 (0.4) &78.9 (0.4) & \textbf{82.0 (0.4)} & \textbf{82.5 (0.5)} & \textbf{82.1 (0.3)} & \textbf{82.5 (0.4)}\\ 
\text{phishing} & 30 &91.1 (0.4) &87.5 (0.3) &92.3 (0.2) & \textbf{93.0 (0.2)} & \textbf{92.7 (0.2)} & \textbf{92.5 (0.3)} & \textbf{92.7 (0.2)}\\ 
\text{waveform} & 21 &86.7 (0.4) &86.2 (0.2) &89.8 (0.3) & \textbf{91.2 (0.3)} & \textbf{91.3 (0.3)} & \textbf{90.7 (0.2)} & \textbf{90.8 (0.3)}\\ 
\text{susy} & 18 &71.3 (0.4) &71.3 (0.6) &74.1 (0.5) & \textbf{77.0 (0.5)} & \textbf{77.5 (0.4)} & \textbf{77.2 (0.3)} & \textbf{77.1 (0.3)}\\ 
\text{w8a} & 300 &87.8 (0.3) &83.6 (0.4) & \textbf{89.6 (0.3)} & \textbf{89.8 (0.3)} &88.2 (0.3) & \textbf{90.2 (0.3)} & \textbf{89.7 (0.3)}\\ 
\text{adult} & 104 &78.8 (0.4) &79.2 (0.3) &78.7 (0.4) & \textbf{80.6 (0.5)} &79.6 (0.4) &79.6 (0.4) & \textbf{80.8 (0.4)}\\ 
\text{twonorm} & 20 &97.2 (0.1) & \textbf{97.7 (0.1)} &97.3 (0.2) & \textbf{97.7 (0.1)} & \textbf{97.5 (0.2)} &97.2 (0.1) &97.2 (0.2)\\ 
\text{mushroom} & 98 &98.3 (0.2) &91.0 (0.5) & \textbf{99.8 (0.0)} & \textbf{99.9 (0.1)} & \textbf{99.8 (0.1)} & \textbf{99.9 (0.0)} & \textbf{99.9 (0.1)}\\ 
\hline
\end{tabular}
\end{table}
\begin{table}[htb]
\centering
\caption{Mean balanced accuracy and standard error for BER minimization from corrupted labels, where $\pi = 0.8$ and $\pi'=0.3$.}
\begin{tabular} { |L|L|L|L|L|L|L|L|L|L|L|L|L|L|L|L|L|L|L|L|L|L|}
\hline
\text{Dataset} & \text{Dim.}& \text{Barrier}& \text{Unhinged}& \text{Sigmoid}& \text{Logistic}& \text{Hinge}& \text{Squared}& \text{Savage}\\ \hline
\text{spambase} & 57 & \textbf{88.3 (0.5)} & \textbf{88.7 (0.3)} & \textbf{88.7 (0.3)} &87.5 (0.4) &87.6 (0.4) &84.4 (0.5) &86.3 (0.5)\\ 
\text{phoneme} & 5 &75.0 (0.5) &75.7 (0.4) &76.9 (0.5) & \textbf{79.3 (0.5)} &79.0 (0.4) & \textbf{79.7 (0.4)} & \textbf{80.2 (0.5)}\\ 
\text{phishing} & 30 &89.9 (0.4) &86.1 (0.4) & \textbf{91.5 (0.3)} &89.7 (0.3) &90.5 (0.3) &85.7 (0.4) &88.5 (0.5)\\ 
\text{waveform} & 21 &87.4 (0.4) &86.8 (0.3) & \textbf{88.7 (0.4)} &87.6 (0.4) & \textbf{88.6 (0.3)} &84.4 (0.5) &87.4 (0.4)\\ 
\text{susy} & 18 &71.1 (0.4) &71.2 (0.5) & \textbf{73.6 (0.4)} & \textbf{73.1 (0.4)} & \textbf{74.1 (0.6)} &71.8 (0.6) & \textbf{73.2 (0.5)}\\ 
\text{w8a} & 300 & \textbf{85.8 (0.5)} &84.0 (0.5) &81.2 (0.4) &76.5 (0.5) &73.2 (0.7) &74.1 (0.5) &78.1 (0.4)\\ 
\text{adult} & 104 & \textbf{77.9 (0.4)} & \textbf{78.1 (0.5)} & \textbf{77.4 (0.4)} &75.2 (0.6) &73.7 (0.5) &70.8 (0.5) &74.6 (0.6)\\ 
\text{twonorm} & 20 & \textbf{97.3 (0.2)} & \textbf{97.6 (0.1)} &97.0 (0.2) &94.3 (0.2) &95.6 (0.2) &89.0 (0.5) &91.8 (0.3)\\ 
\text{mushroom} & 98 &97.9 (0.3) &94.8 (0.6) & \textbf{99.1 (0.2)} &97.5 (0.2) & \textbf{98.9 (0.1)} &93.6 (0.3) &97.7 (0.2)\\ 
\hline
\end{tabular}
\end{table}
\begin{table}[htb]
\centering
\caption{Mean balanced accuracy and standard error for BER minimization from corrupted labels, where $\pi = 0.7$ and $\pi'=0.4$.}
\begin{tabular} { |L|L|L|L|L|L|L|L|L|L|L|L|L|L|L|L|L|L|L|L|L|L|}
\hline
\text{Dataset} & \text{Dim.}& \text{Barrier}& \text{Unhinged}& \text{Sigmoid}& \text{Logistic}& \text{Hinge}& \text{Squared}& \text{Savage}\\ \hline
\text{spambase} & 57 &85.6 (0.4) & \textbf{87.6 (0.3)} &86.1 (0.4) &81.7 (0.5) &80.4 (0.6) &76.1 (0.5) &79.4 (0.5)\\ 
\text{phoneme} & 5 & \textbf{75.8 (0.3)} &75.5 (0.6) & \textbf{76.8 (0.7)} & \textbf{76.9 (0.6)} & \textbf{76.1 (0.6)} & \textbf{76.6 (0.8)} & \textbf{76.2 (0.7)}\\ 
\text{phishing} & 30 & \textbf{87.9 (0.7)} &86.0 (0.5) & \textbf{89.2 (0.5)} &84.1 (0.5) &84.4 (0.6) &77.5 (0.5) &82.2 (0.6)\\ 
\text{waveform} & 21 &86.6 (0.3) &86.6 (0.5) & \textbf{88.3 (0.4)} &82.4 (0.4) &84.6 (0.5) &76.0 (0.6) &79.4 (0.6)\\ 
\text{susy} & 18 &70.2 (0.5) & \textbf{70.6 (0.7)} & \textbf{71.3 (0.4)} &68.3 (0.8) &68.4 (0.5) &66.9 (0.5) &67.8 (0.5)\\ 
\text{w8a} & 300 &77.7 (0.7) & \textbf{80.4 (0.6)} &71.2 (0.6) &68.0 (0.5) &65.9 (0.6) &65.7 (0.6) &68.4 (0.8)\\ 
\text{adult} & 104 & \textbf{75.9 (0.4)} & \textbf{76.9 (0.6)} &75.3 (0.5) &69.4 (0.5) &69.0 (0.6) &63.2 (0.6) &67.4 (0.5)\\ 
\text{twonorm} & 20 &96.7 (0.2) & \textbf{97.2 (0.1)} &96.4 (0.2) &86.7 (0.4) &90.1 (0.4) &78.8 (0.6) &83.7 (0.4)\\ 
\text{mushroom} & 98 & \textbf{96.8 (0.5)} &92.2 (0.9) & \textbf{96.6 (0.5)} &90.8 (0.5) &95.1 (0.6) &79.5 (0.6) &90.2 (0.4)\\ 
\hline
\end{tabular}
\end{table}
\begin{table}[htb]
\centering
\caption{Mean balanced accuracy and standard error for BER minimization from corrupted labels, where $\pi = 0.65$ and $\pi'=0.45$.}
\begin{tabular} { |L|L|L|L|L|L|L|L|L|L|L|L|L|L|L|L|L|L|L|L|L|L|}
\hline
\text{Dataset} & \text{Dim.}& \text{Barrier}& \text{Unhinged}& \text{Sigmoid}& \text{Logistic}& \text{Hinge}& \text{Squared}& \text{Savage}\\ \hline
\text{spambase} & 57 &82.3 (0.8) & \textbf{84.1 (0.6)} &80.9 (0.6) &72.6 (0.7) &74.7 (0.7) &69.5 (0.7) &73.6 (0.6)\\ 
\text{phoneme} & 5 & \textbf{74.5 (0.8)} & \textbf{73.4 (0.9)} & \textbf{74.5 (0.6)} & \textbf{73.4 (0.8)} & \textbf{73.8 (1.1)} &71.3 (0.9) &71.0 (0.7)\\ 
\text{phishing} & 30 & \textbf{86.2 (0.4)} &82.8 (0.7) &84.9 (0.7) &77.7 (0.6) &78.8 (0.9) &69.1 (0.8) &73.3 (0.7)\\ 
\text{waveform} & 21 & \textbf{86.1 (0.4)} & \textbf{87.1 (0.6)} &85.4 (0.6) &75.8 (0.7) &78.3 (0.7) &69.2 (0.6) &73.2 (0.6)\\ 
\text{susy} & 18 & \textbf{68.3 (0.6)} & \textbf{68.9 (0.8)} & \textbf{66.9 (0.9)} &64.8 (0.8) &65.1 (0.8) &61.7 (0.7) &64.6 (0.7)\\ 
\text{w8a} & 300 &71.3 (0.8) & \textbf{73.1 (0.5)} &65.1 (0.7) &62.4 (0.7) &61.1 (0.6) &60.6 (0.5) &62.3 (0.6)\\ 
\text{adult} & 104 &73.2 (0.7) & \textbf{74.7 (0.6)} &69.9 (1.0) &64.8 (0.8) &64.2 (1.0) &59.1 (0.6) &63.2 (0.8)\\ 
\text{twonorm} & 20 & \textbf{96.2 (0.3)} & \textbf{96.7 (0.2)} &95.4 (0.4) &80.2 (0.5) &82.8 (0.9) &71.6 (0.7) &75.9 (0.6)\\ 
\text{mushroom} & 98 & \textbf{93.4 (0.8)} &91.1 (0.9) & \textbf{94.4 (0.7)} &81.3 (0.5) &84.5 (1.0) &72.2 (0.6) &79.5 (0.8)\\ 
\hline
\end{tabular}
\end{table}

\FloatBarrier
\subsection{AUC Optimization Using UCI and LIBSVM Datasets}
Outperforming methods are highlighted in boldface using one-sided t-test with the significance level 5\%. The experiments were conducted 20 times.
\begin{table}[htb]
\centering
\caption{Mean AUC score and standard error for AUC maximization from corrupted labels, where $\pi = 1.0$ and $\pi'=0.0$.}
\begin{tabular} { |L|L|L|L|L|L|L|L|L|L|L|L|L|L|L|L|L|L|L|L|L|L|}
\hline
\text{Dataset} & \text{Dim.}& \text{Barrier}& \text{Unhinged}& \text{Sigmoid}& \text{Logistic}& \text{Hinge}& \text{Squared}& \text{Savage}\\ \hline
\text{spambase} & 57 &94.4 (0.3) &93.7 (0.2) &95.9 (0.1) &96.4 (0.2) & \textbf{97.0 (0.2)} & \textbf{96.8 (0.2)} &96.5 (0.2)\\ 
\text{phoneme} & 5 &81.8 (0.5) &82.3 (0.4) &84.2 (0.3) & \textbf{87.4 (0.3)} & \textbf{88.1 (0.4)} & \textbf{87.3 (0.3)} & \textbf{87.9 (0.4)}\\ 
\text{phishing} & 30 &97.3 (0.1) &93.9 (0.2) &97.6 (0.1) & \textbf{97.9 (0.1)} & \textbf{97.9 (0.1)} & \textbf{97.7 (0.1)} & \textbf{97.8 (0.1)}\\ 
\text{waveform} & 21 &95.3 (0.2) &90.3 (0.4) &96.0 (0.2) & \textbf{96.3 (0.2)} & \textbf{96.8 (0.1)} &96.1 (0.2) & \textbf{96.6 (0.1)}\\ 
\text{susy} & 18 &81.3 (0.3) &78.1 (0.6) &83.1 (0.5) & \textbf{84.7 (0.4)} & \textbf{85.5 (0.4)} & \textbf{85.0 (0.4)} &84.5 (0.3)\\ 
\text{w8a} & 300 &96.5 (0.2) &94.5 (0.2) & \textbf{96.9 (0.2)} & \textbf{96.8 (0.1)} &96.7 (0.1) & \textbf{96.7 (0.2)} & \textbf{97.1 (0.1)}\\ 
\text{adult} & 104 &86.1 (0.3) &87.6 (0.2) &87.4 (0.3) & \textbf{88.6 (0.4)} & \textbf{88.3 (0.3)} &87.6 (0.3) & \textbf{88.8 (0.3)}\\ 
\text{twonorm} & 20 &99.7 (0.0) & \textbf{99.8 (0.0)} &99.7 (0.0) & \textbf{99.8 (0.0)} &99.7 (0.0) &99.6 (0.0) &99.7 (0.0)\\ 
\text{mushroom} & 98 & \textbf{99.9 (0.0)} &99.6 (0.1) & \textbf{100.0 (0.0)} & \textbf{100.0 (0.0)} & \textbf{100.0 (0.0)} & \textbf{100.0 (0.0)} & \textbf{99.9 (0.1)}\\ 
\hline
\end{tabular}
\end{table}
\begin{table}[htb]
\centering
\caption{Mean AUC score and standard error for AUC maximization from corrupted labels, where $\pi = 0.8$ and $\pi'=0.3$.}
\begin{tabular} { |L|L|L|L|L|L|L|L|L|L|L|L|L|L|L|L|L|L|L|L|L|L|}
\hline
\text{Dataset} & \text{Dim.}& \text{Barrier}& \text{Unhinged}& \text{Sigmoid}& \text{Logistic}& \text{Hinge}& \text{Squared}& \text{Savage}\\ \hline
\text{spambase} & 57 & \textbf{93.8 (0.3)} & \textbf{94.3 (0.2)} & \textbf{94.1 (0.3)} &93.6 (0.3) &92.3 (0.3) &90.5 (0.5) &92.7 (0.5)\\ 
\text{phoneme} & 5 &81.0 (0.5) &81.7 (0.4) &82.1 (0.5) & \textbf{85.3 (0.4)} & \textbf{85.1 (0.2)} & \textbf{85.6 (0.3)} & \textbf{85.7 (0.4)}\\ 
\text{phishing} & 30 & \textbf{96.8 (0.1)} &93.7 (0.3) & \textbf{96.8 (0.2)} &96.2 (0.2) &95.1 (0.2) &92.9 (0.3) &95.4 (0.2)\\ 
\text{waveform} & 21 & \textbf{94.7 (0.2)} &91.3 (0.3) & \textbf{95.1 (0.3)} &94.1 (0.3) &93.8 (0.2) &91.5 (0.4) &94.1 (0.3)\\ 
\text{susy} & 18 &80.0 (0.5) &77.9 (0.5) & \textbf{81.3 (0.4)} & \textbf{81.1 (0.4)} & \textbf{81.7 (0.5)} &79.0 (0.6) & \textbf{80.8 (0.5)}\\ 
\text{w8a} & 300 &91.3 (0.5) & \textbf{92.9 (0.2)} &90.8 (0.3) &87.4 (0.4) &83.2 (0.6) &82.9 (0.6) &88.7 (0.4)\\ 
\text{adult} & 104 &85.3 (0.3) & \textbf{86.1 (0.4)} &85.1 (0.4) &82.2 (0.5) &78.3 (0.6) &77.4 (0.5) &81.9 (0.5)\\ 
\text{twonorm} & 20 &99.7 (0.0) & \textbf{99.8 (0.0)} &99.4 (0.0) &98.9 (0.1) &98.3 (0.1) &95.1 (0.2) &97.5 (0.1)\\ 
\text{mushroom} & 98 & \textbf{99.8 (0.1)} &99.3 (0.1) & \textbf{99.7 (0.1)} &99.2 (0.2) &98.6 (0.2) &97.9 (0.2) &99.6 (0.1)\\ 
\hline
\end{tabular}
\end{table}
\begin{table}[htb]
\centering
\caption{Mean AUC score and standard error for AUC maximization from corrupted labels, where $\pi = 0.7$ and $\pi'=0.4$.}
\begin{tabular} { |L|L|L|L|L|L|L|L|L|L|L|L|L|L|L|L|L|L|L|L|L|L|}
\hline
\text{Dataset} & \text{Dim.}& \text{Barrier}& \text{Unhinged}& \text{Sigmoid}& \text{Logistic}& \text{Hinge}& \text{Squared}& \text{Savage}\\ \hline
\text{spambase} & 57 &90.4 (0.4) & \textbf{93.4 (0.3)} &91.8 (0.3) &88.3 (0.5) &85.4 (0.6) &82.2 (0.6) &86.0 (0.5)\\ 
\text{phoneme} & 5 & \textbf{81.0 (0.4)} & \textbf{81.1 (0.5)} & \textbf{82.2 (0.6)} & \textbf{82.2 (0.6)} & \textbf{81.8 (0.5)} & \textbf{82.2 (0.6)} & \textbf{81.9 (0.6)}\\ 
\text{phishing} & 30 & \textbf{95.9 (0.3)} &93.0 (0.5) &94.9 (0.4) &91.7 (0.4) &88.1 (0.5) &83.7 (0.5) &90.2 (0.5)\\ 
\text{waveform} & 21 & \textbf{93.5 (0.3)} &91.5 (0.5) & \textbf{94.1 (0.2)} &90.1 (0.4) &88.6 (0.6) &82.4 (0.8) &86.5 (0.5)\\ 
\text{susy} & 18 & \textbf{77.9 (0.6)} &77.4 (0.6) & \textbf{78.8 (0.5)} &75.6 (1.0) &74.6 (0.6) &73.2 (0.6) &74.2 (0.7)\\ 
\text{w8a} & 300 &79.1 (0.7) & \textbf{89.5 (0.5)} &79.4 (0.6) &75.6 (0.4) &72.3 (0.8) &71.5 (0.6) &76.4 (0.8)\\ 
\text{adult} & 104 &82.1 (0.4) & \textbf{84.6 (0.4)} &81.7 (0.5) &75.5 (0.5) &72.6 (0.6) &68.2 (0.8) &73.4 (0.6)\\ 
\text{twonorm} & 20 &99.4 (0.1) & \textbf{99.7 (0.0)} &98.9 (0.1) &94.5 (0.3) &92.3 (0.5) &85.4 (0.6) &91.6 (0.3)\\ 
\text{mushroom} & 98 & \textbf{99.6 (0.1)} &98.9 (0.1) &98.8 (0.2) &96.6 (0.3) &92.6 (0.5) &86.7 (0.5) &96.7 (0.3)\\ 
\hline
\end{tabular}
\end{table}
\begin{table}[htb]
\centering
\caption{Mean AUC score and standard error for AUC maximization from corrupted labels, where $\pi = 0.65$ and $\pi'=0.45$.}
\begin{tabular} { |L|L|L|L|L|L|L|L|L|L|L|L|L|L|L|L|L|L|L|L|L|L|}
\hline
\text{Dataset} & \text{Dim.}& \text{Barrier}& \text{Unhinged}& \text{Sigmoid}& \text{Logistic}& \text{Hinge}& \text{Squared}& \text{Savage}\\ \hline
\text{spambase} & 57 &86.8 (0.7) & \textbf{90.9 (0.4)} &86.0 (0.4) &79.2 (0.8) &77.7 (0.7) &73.6 (0.8) &80.1 (0.8)\\ 
\text{phoneme} & 5 & \textbf{80.2 (0.6)} & \textbf{79.2 (0.9)} &78.4 (0.8) &78.2 (0.8) &77.8 (0.8) &76.2 (0.8) &76.2 (0.7)\\ 
\text{phishing} & 30 & \textbf{94.7 (0.3)} &90.2 (0.8) &91.1 (0.6) &85.0 (0.6) &82.0 (0.8) &73.8 (0.9) &80.3 (0.8)\\ 
\text{waveform} & 21 & \textbf{92.2 (0.4)} & \textbf{91.7 (0.6)} & \textbf{90.9 (0.6)} &82.3 (0.7) &79.8 (0.9) &75.1 (0.7) &80.1 (0.6)\\ 
\text{susy} & 18 & \textbf{73.6 (0.8)} & \textbf{75.3 (0.8)} &72.5 (1.0) &70.9 (1.0) &69.9 (1.0) &66.2 (0.8) &69.9 (0.9)\\ 
\text{w8a} & 300 &70.9 (0.8) & \textbf{81.7 (0.8)} &71.3 (0.9) &68.4 (0.7) &66.8 (0.8) &65.5 (0.6) &68.3 (0.6)\\ 
\text{adult} & 104 &79.0 (0.7) & \textbf{81.2 (0.7)} &75.3 (1.1) &69.6 (0.8) &66.8 (1.0) &62.3 (0.8) &68.0 (1.0)\\ 
\text{twonorm} & 20 &99.1 (0.1) & \textbf{99.6 (0.0)} &98.0 (0.2) &88.3 (0.5) &83.9 (0.7) &77.3 (0.7) &82.7 (0.5)\\ 
\text{mushroom} & 98 & \textbf{98.4 (0.2)} &97.2 (0.4) & \textbf{97.8 (0.3)} &89.0 (0.5) &82.2 (0.6) &77.8 (0.6) &88.1 (0.7)\\ 
\hline
\end{tabular}
\end{table}

\FloatBarrier
\subsection{BER Minimization Using MNIST Dataset}
Outperforming methods are highlighted in boldface using one-sided t-test with the significance level 5\%. The experiments were conducted 10 times.
\begin{table}[htb]
\centering
\caption{Mean balanced accuracy and standard error for BER minimization from corrupted labels with varying noises.}
\begin{tabular}{|L|C|L|L | L | L | L|L|L|  }
\hline
\text{Dataset} & (\pi,\pi') &\text{Barrier}& \text{Unhinged}& \text{Sigmoid}& \text{Logistic}& \text{Hinge}& \text{Squared}& \text{Savage}\\ \hline
\multirow{4}{*}{\text{MNIST} }
	& (1.0,0.0)  &97.8 (0.0) &50.2 (0.1) &99.0 (0.0) & \textbf{99.1 (0.0)} &99.0 (0.0) &98.4 (0.0) &99.0 (0.0)\\ 
	{} & (0.8,0.3) & \textbf{97.3 (0.0)} &50.4 (0.2) &96.7 (0.1) &80.5 (0.2) &80.4 (0.2) &89.9 (0.4) &81.7 (0.8)\\ 
	{} & (0.7,0.4) & \textbf{95.8 (0.2)} &50.0 (0.0) &92.7 (0.3) &69.6 (0.3) &69.5 (0.2) &81.8 (1.2) &70.2 (0.9)\\ 
	{} & (0.65,0.45) & \textbf{92.8 (0.3)} &50.0 (0.0) &83.1 (3.7) &64.0 (0.2) &63.7 (0.3) &73.0 (1.3) &63.9 (0.1)\\
 \hline
\end{tabular}
\end{table}

\FloatBarrier
\subsection{AUC Maximization Using MNIST Dataset}
Outperforming methods are highlighted in boldface using one-sided t-test with the significance level 5\%. The experiments were conducted 10 times.
\begin{table}[htb]
\centering
\caption{Mean AUC score and standard error for AUC maximization from corrupted labels with varying noises.}
\begin{tabular}{|L|C|L|L | L | L | L|L|L|  }
\hline
\text{Dataset} & (\pi,\pi') &\text{Barrier}& \text{Unhinged}& \text{Sigmoid}& \text{Logistic}& \text{Hinge}& \text{Squared}& \text{Savage}\\ \hline
\multirow{4}{*}{\text{MNIST} }
	& (1.0,0.0) &99.6 (0.0) &85.0 (0.5) &99.8 (0.0) & \textbf{99.8 (0.0)} &99.8 (0.0) &99.5 (0.0) &99.7 (0.0)\\ 
	{} & (0.8,0.3)& \textbf{99.4 (0.0)} &84.3 (0.4) &98.0 (0.1) &88.5 (0.3) &88.2 (0.2) &96.6 (0.2) &97.2 (0.4)\\  
	{} & (0.7,0.4) & \textbf{99.0 (0.0)} &83.1 (0.4) &95.9 (0.2) &75.5 (0.4) &76.1 (0.3) &87.5 (0.6) &94.7 (0.4)\\  
	{} & (0.65,0.45)& \textbf{96.9 (0.2)} &80.6 (0.3) &92.2 (0.7) &68.6 (0.4) &68.5 (0.4) &80.2 (0.5) &90.6 (1.0)\\ 
 \hline
\end{tabular}
\end{table}
\FloatBarrier
\subsection{BER Minimization Using CIFAR-10 Dataset}
Outperforming methods are highlighted in boldface using one-sided t-test with the significance level 5\%. The experiments were conducted 10 times.
\begin{table}[htb]
\centering
\caption{Mean balanced accuracy and standard error for BER minimization from corrupted labels, where $\pi = 1.0$ $\pi' = 0.0$}
\begin{tabular} { |L|L|L|L|L|L|L|L|L|L|L|L|L|L|L|L|L|L|}
\hline
\text{Dataset}& \text{Barrier}& \text{Unhinged}& \text{Sigmoid}& \text{Logistic}& \text{Hinge}& \text{Squared}& \text{Savage}\\ \hline
\text{automobile} &87.0 (0.4) &69.5 (0.2) &93.0 (0.2) &93.6 (0.1) &93.4 (0.1) & \textbf{94.3 (0.1)} &93.4 (0.1)\\ 
\text{bird} &84.0 (0.2) &64.9 (0.1) &88.2 (0.1) & \textbf{88.6 (0.2)} & \textbf{88.7 (0.2)} & \textbf{88.9 (0.1)} &88.6 (0.1)\\ 
\text{car} &88.5 (0.1) &69.8 (0.1) &91.8 (0.1) &92.5 (0.2) &92.6 (0.1) & \textbf{93.1 (0.1)} &92.8 (0.1)\\ 
\text{deer} &89.6 (0.1) &71.6 (0.2) &93.3 (0.1) &93.7 (0.2) &93.8 (0.0) & \textbf{94.1 (0.1)} & \textbf{94.0 (0.1)}\\ 
\text{dog} &91.6 (0.1) &67.6 (0.2) &93.8 (0.1) &94.1 (0.2) &94.2 (0.1) & \textbf{94.9 (0.1)} &94.4 (0.1)\\ 
\text{frog} &93.3 (0.1) &73.8 (0.1) &95.6 (0.1) & \textbf{96.2 (0.1)} & \textbf{96.0 (0.1)} & \textbf{96.1 (0.1)} &96.0 (0.1)\\ 
\text{horse} &92.8 (0.1) &69.0 (0.2) &94.5 (0.1) &94.9 (0.1) &94.6 (0.1) & \textbf{95.3 (0.1)} &94.9 (0.1)\\ 
\text{ship} &80.9 (0.5) &64.4 (0.1) &87.5 (0.3) &89.1 (0.2) &89.1 (0.2) & \textbf{89.6 (0.1)} &89.3 (0.1)\\ 
\text{truck} &87.5 (0.2) &69.4 (0.1) &90.6 (0.2) &91.2 (0.2) &91.1 (0.2) & \textbf{91.6 (0.1)} &91.1 (0.2)\\ 
\hline
\end{tabular}
\end{table}
\begin{table}[htb]
\centering
\caption{Mean balanced accuracy and standard error for BER minimization from corrupted labels, where $\pi = 0.8$ $\pi' = 0.3$}
\begin{tabular} { |L|L|L|L|L|L|L|L|L|L|L|L|L|L|L|L|L|L|}
\hline
\text{Dataset}& \text{Barrier}& \text{Unhinged}& \text{Sigmoid}& \text{Logistic}& \text{Hinge}& \text{Squared}& \text{Savage}\\ \hline
\text{automobile} &86.6 (0.2) &70.2 (0.2) & \textbf{88.5 (0.2)} &74.3 (0.2) &74.1 (0.5) &74.2 (0.3) &73.6 (0.3)\\ 
\text{bird} &82.3 (0.3) &66.7 (0.2) & \textbf{83.3 (0.3)} &72.4 (0.4) &72.3 (0.4) &71.6 (0.3) &71.3 (0.4)\\ 
\text{car} &87.3 (0.1) &71.2 (0.1) & \textbf{87.8 (0.1)} &73.3 (0.2) &74.4 (0.4) &73.9 (0.3) &73.5 (0.4)\\ 
\text{deer} & \textbf{88.5 (0.2)} &72.9 (0.2) & \textbf{88.9 (0.1)} &74.3 (0.4) &75.3 (0.5) &74.6 (0.4) &74.2 (0.3)\\ 
\text{dog} &90.0 (0.1) &68.4 (0.1) & \textbf{90.6 (0.2)} &75.4 (0.4) &76.6 (0.4) &75.9 (0.2) &74.4 (0.5)\\ 
\text{frog} &92.8 (0.1) &76.2 (0.2) & \textbf{93.1 (0.1)} &76.0 (0.3) &78.2 (0.6) &77.9 (0.3) &76.5 (0.5)\\ 
\text{horse} & \textbf{90.8 (0.3)} &71.0 (0.1) &89.8 (0.2) &76.0 (0.3) &76.8 (0.4) &76.3 (0.3) &75.5 (0.3)\\ 
\text{ship} &77.1 (0.3) &65.7 (0.1) & \textbf{80.0 (0.2)} &70.1 (0.2) &69.7 (0.2) &69.8 (0.3) &69.8 (0.3)\\ 
\text{truck} & \textbf{86.3 (0.1)} &70.0 (0.1) & \textbf{86.3 (0.3)} &73.9 (0.3) &73.8 (0.5) &74.6 (0.3) &73.7 (0.4)\\ 
\hline
\end{tabular}
\end{table}
\begin{table}[htb]
\centering
\caption{Mean balanced accuracy and standard error for BER minimization from corrupted labels, where $\pi = 0.7$ $\pi' = 0.4$}
\begin{tabular} { |L|L|L|L|L|L|L|L|L|L|L|L|L|L|L|L|L|L|}
\hline
\text{Dataset}& \text{Barrier}& \text{Unhinged}& \text{Sigmoid}& \text{Logistic}& \text{Hinge}& \text{Squared}& \text{Savage}\\ \hline
\text{automobile} & \textbf{85.4 (0.2)} &70.0 (0.3) &83.8 (0.3) &65.8 (0.5) &65.6 (0.4) &65.6 (0.3) &64.1 (0.4)\\ 
\text{bird} & \textbf{81.7 (0.2)} &66.9 (0.1) &80.7 (0.3) &63.1 (0.4) &63.6 (0.5) &63.6 (0.3) &62.8 (0.4)\\ 
\text{car} & \textbf{86.7 (0.2)} &71.4 (0.1) &84.3 (0.2) &64.8 (0.5) &64.4 (0.4) &64.5 (0.2) &64.8 (0.4)\\ 
\text{deer} & \textbf{87.5 (0.1)} &74.0 (0.1) &84.3 (0.2) &64.0 (0.5) &63.8 (0.6) &64.5 (0.2) &64.1 (0.5)\\ 
\text{dog} & \textbf{88.9 (0.2)} &68.4 (0.1) &87.2 (0.2) &65.8 (0.7) &64.5 (0.5) &65.2 (0.3) &64.9 (0.4)\\ 
\text{frog} & \textbf{92.3 (0.1)} &77.0 (0.2) &90.9 (0.2) &65.6 (0.7) &66.0 (0.5) &66.6 (0.4) &67.0 (0.4)\\ 
\text{horse} & \textbf{88.8 (0.2)} &71.2 (0.2) &86.1 (0.3) &65.7 (0.6) &65.6 (0.4) &66.0 (0.4) &65.5 (0.3)\\ 
\text{ship} & \textbf{74.9 (0.2)} &65.5 (0.0) & \textbf{74.7 (0.2)} &62.1 (0.4) &61.4 (0.5) &62.2 (0.4) &62.4 (0.2)\\ 
\text{truck} & \textbf{84.7 (0.2)} &70.3 (0.1) &82.4 (0.3) &63.7 (0.3) &64.4 (0.6) &64.5 (0.4) &64.3 (0.5)\\ 
\hline
\end{tabular}
\end{table}
\begin{table}[htb]
\centering
\caption{Mean balanced accuracy and standard error for BER minimization from corrupted labels, where $\pi = 0.65$ $\pi' = 0.45$}
\begin{tabular} { |L|L|L|L|L|L|L|L|L|L|L|L|L|L|L|L|L|L|}
\hline
\text{Dataset}& \text{Barrier}& \text{Unhinged}& \text{Sigmoid}& \text{Logistic}& \text{Hinge}& \text{Squared}& \text{Savage}\\ \hline
\text{automobile} & \textbf{84.0 (0.3)} &70.8 (0.2) &79.7 (0.3) &59.8 (0.6) &59.1 (0.5) &60.1 (0.3) &60.4 (0.3)\\ 
\text{bird} & \textbf{81.5 (0.2)} &67.6 (0.2) &77.5 (0.4) &58.5 (0.4) &59.4 (0.3) &58.4 (0.4) &58.0 (0.5)\\ 
\text{car} & \textbf{85.6 (0.1)} &71.8 (0.1) &81.5 (0.4) &60.6 (0.4) &59.7 (0.4) &59.8 (0.3) &60.1 (0.2)\\ 
\text{deer} & \textbf{86.2 (0.2)} &74.6 (0.2) &80.3 (0.5) &58.3 (0.4) &58.9 (0.5) &59.0 (0.4) &58.5 (0.4)\\ 
\text{dog} & \textbf{87.2 (0.4)} &68.6 (0.2) &83.1 (0.2) &59.7 (0.2) &59.8 (0.6) &59.9 (0.4) &59.3 (0.4)\\ 
\text{frog} & \textbf{91.0 (0.2)} &78.2 (0.1) &88.6 (0.3) &60.4 (0.5) &61.0 (0.4) &60.9 (0.3) &61.6 (0.5)\\ 
\text{horse} & \textbf{86.4 (0.4)} &71.4 (0.1) &82.6 (0.3) &60.3 (0.4) &60.0 (0.4) &60.0 (0.4) &60.0 (0.3)\\ 
\text{ship} & \textbf{71.7 (0.6)} &65.9 (0.1) &68.9 (0.4) &58.2 (0.3) &58.4 (0.3) &57.2 (0.3) &58.1 (0.3)\\ 
\text{truck} & \textbf{82.4 (0.2)} &70.6 (0.1) &78.6 (0.4) &60.0 (0.4) &59.1 (0.4) &60.0 (0.2) &59.5 (0.4)\\ 
\hline
\end{tabular}
\end{table}

\FloatBarrier

\subsection{AUC Maximization Using CIFAR-10 Dataset}
Outperforming methods are highlighted in boldface using one-sided t-test with the significance level 5\%. The experiments were conducted 10 times.
\begin{table}[htb]
\centering
\caption{Mean AUC score and standard error for AUC maximization from corrupted labels, $\pi = 1.0$ $\pi' = 0.0$}
\begin{tabular} { |L|L|L|L|L|L|L|L|L|L|L|L|L|L|L|L|L|L|}
\hline
\text{Dataset}& \text{Barrier}& \text{Unhinged}& \text{Sigmoid}& \text{Logistic}& \text{Hinge}& \text{Squared}& \text{Savage}\\ \hline
\text{automobile} &95.8 (0.1) &75.2 (0.1) & \textbf{98.4 (0.0)} & \textbf{98.4 (0.0)} & \textbf{98.3 (0.0)} & \textbf{98.3 (0.1)} &98.2 (0.0)\\ 
\text{bird} &91.5 (0.1) &71.7 (0.0) &95.0 (0.1) & \textbf{95.2 (0.0)} & \textbf{95.2 (0.0)} & \textbf{95.2 (0.1)} & \textbf{95.1 (0.1)}\\ 
\text{car} &94.7 (0.1) &76.5 (0.0) &97.5 (0.1) &97.6 (0.0) & \textbf{97.6 (0.1)} & \textbf{97.7 (0.0)} & \textbf{97.7 (0.0)}\\ 
\text{deer} &95.3 (0.1) &79.5 (0.1) & \textbf{98.3 (0.1)} & \textbf{98.3 (0.1)} & \textbf{98.4 (0.0)} & \textbf{98.3 (0.1)} & \textbf{98.3 (0.1)}\\ 
\text{dog} &96.5 (0.1) &74.4 (0.2) & \textbf{98.5 (0.0)} & \textbf{98.5 (0.0)} & \textbf{98.5 (0.0)} & \textbf{98.5 (0.0)} & \textbf{98.5 (0.1)}\\ 
\text{frog} &97.2 (0.1) &81.5 (0.0) &99.1 (0.0) &99.0 (0.0) & \textbf{99.1 (0.0)} &99.0 (0.0) &99.0 (0.0)\\ 
\text{horse} &97.2 (0.1) &76.1 (0.0) & \textbf{98.9 (0.0)} & \textbf{98.9 (0.0)} & \textbf{98.9 (0.0)} &98.7 (0.0) & \textbf{98.8 (0.0)}\\ 
\text{ship} &92.4 (0.1) &70.7 (0.1) &95.5 (0.1) & \textbf{95.7 (0.1)} &95.5 (0.1) & \textbf{95.5 (0.1)} & \textbf{95.6 (0.1)}\\ 
\text{truck} &94.2 (0.1) &75.7 (0.1) & \textbf{97.2 (0.0)} & \textbf{97.1 (0.0)} & \textbf{97.1 (0.1)} &96.9 (0.1) &97.1 (0.0)\\ 
\hline
\end{tabular}
\end{table}
\begin{table}[htb]
\centering
\caption{Mean AUC score and standard error for AUC maximization from corrupted labels, $\pi = 0.8$ $\pi' = 0.3$}
\begin{tabular} { |L|L|L|L|L|L|L|L|L|L|L|L|L|L|L|L|L|L|}
\hline
\text{Dataset}& \text{Barrier}& \text{Unhinged}& \text{Sigmoid}& \text{Logistic}& \text{Hinge}& \text{Squared}& \text{Savage}\\ \hline
\text{automobile} & \textbf{94.8 (0.1)} &75.7 (0.1) &83.5 (0.5) &82.0 (0.4) &81.4 (0.2) &82.1 (0.3) &82.1 (0.3)\\ 
\text{bird} & \textbf{90.6 (0.1)} &73.3 (0.0) &79.8 (0.3) &79.2 (0.3) &78.7 (0.3) &79.1 (0.2) &79.4 (0.3)\\ 
\text{car} & \textbf{93.4 (0.4)} &78.4 (0.0) &82.7 (0.5) &82.1 (0.3) &81.1 (0.3) &82.0 (0.4) &80.8 (0.4)\\ 
\text{deer} & \textbf{94.6 (0.1)} &81.2 (0.0) &83.6 (0.6) &81.5 (0.5) &81.6 (0.3) &82.1 (0.4) &82.3 (0.3)\\ 
\text{dog} & \textbf{95.6 (0.1)} &76.2 (0.1) &82.8 (0.4) &83.5 (0.4) &82.8 (0.4) &83.3 (0.3) &83.1 (0.3)\\ 
\text{frog} & \textbf{96.9 (0.1)} &83.7 (0.0) &85.2 (0.4) &85.2 (0.4) &84.4 (0.2) &84.6 (0.4) &84.3 (0.4)\\ 
\text{horse} & \textbf{96.2 (0.4)} &78.1 (0.1) &84.0 (0.5) &84.3 (0.3) &83.9 (0.5) &83.9 (0.3) &83.9 (0.4)\\ 
\text{ship} & \textbf{89.0 (0.1)} &71.9 (0.1) &78.3 (0.4) &76.8 (0.3) &77.4 (0.3) &77.0 (0.4) &77.2 (0.3)\\ 
\text{truck} & \textbf{93.7 (0.1)} &76.6 (0.1) &81.4 (0.7) &81.7 (0.3) &81.1 (0.2) &81.0 (0.4) &81.9 (0.3)\\ 
\hline
\end{tabular}
\end{table}
\begin{table}[htb]
\centering
\caption{Mean AUC score and standard error for AUC maximization from corrupted labels, $\pi = 0.7$ $\pi' = 0.4$}
\begin{tabular} { |L|L|L|L|L|L|L|L|L|L|L|L|L|L|L|L|L|L|}
\hline
\text{Dataset}& \text{Barrier}& \text{Unhinged}& \text{Sigmoid}& \text{Logistic}& \text{Hinge}& \text{Squared}& \text{Savage}\\ \hline
\text{automobile} & \textbf{93.2 (0.1)} &76.0 (0.1) &72.3 (0.8) &71.1 (0.6) &70.3 (0.3) &71.0 (0.5) &70.7 (0.4)\\ 
\text{bird} & \textbf{90.0 (0.2)} &73.8 (0.0) &68.7 (0.8) &68.2 (0.3) &68.3 (0.5) &67.0 (0.4) &67.9 (0.5)\\ 
\text{car} & \textbf{93.4 (0.2)} &78.5 (0.0) &70.5 (0.4) &70.2 (0.5) &69.2 (0.5) &70.0 (0.5) &69.8 (0.2)\\ 
\text{deer} & \textbf{93.3 (0.2)} &81.6 (0.1) &69.3 (0.6) &69.5 (0.7) &69.5 (0.4) &69.3 (0.3) &69.9 (0.5)\\ 
\text{dog} & \textbf{94.9 (0.1)} &76.4 (0.1) &70.9 (0.8) &71.8 (0.4) &70.9 (0.4) &70.9 (0.4) &71.5 (0.3)\\ 
\text{frog} & \textbf{96.7 (0.1)} &84.8 (0.0) &73.1 (0.7) &73.4 (0.4) &72.9 (0.6) &72.3 (0.4) &72.7 (0.4)\\ 
\text{horse} & \textbf{95.8 (0.1)} &78.4 (0.1) &72.3 (0.7) &72.6 (0.4) &70.8 (0.3) &71.2 (0.5) &71.7 (0.4)\\ 
\text{ship} & \textbf{84.5 (0.4)} &71.6 (0.1) &69.8 (0.4) &67.2 (0.4) &66.4 (0.3) &66.9 (0.4) &67.4 (0.3)\\ 
\text{truck} & \textbf{92.1 (0.1)} &76.8 (0.1) &71.3 (0.7) &70.1 (0.4) &69.2 (0.5) &69.8 (0.4) &70.3 (0.3)\\ 
\hline
\end{tabular}
\end{table}
\begin{table}[htb]
\centering
\caption{Mean AUC score and standard error for AUC maximization from corrupted labels, $\pi = 0.65$ $\pi' = 0.45$}
\begin{tabular} { |L|L|L|L|L|L|L|L|L|L|L|L|L|L|L|L|L|L|}
\hline
\text{Dataset}& \text{Barrier}& \text{Unhinged}& \text{Sigmoid}& \text{Logistic}& \text{Hinge}& \text{Squared}& \text{Savage}\\ \hline
\text{automobile} & \textbf{91.3 (0.3)} &76.5 (0.1) &64.1 (0.4) &64.7 (0.3) &63.9 (0.3) &64.0 (0.6) &64.0 (0.5)\\ 
\text{bird} & \textbf{88.5 (0.1)} &74.4 (0.0) &63.3 (0.6) &62.0 (0.5) &61.4 (0.4) &61.6 (0.3) &62.0 (0.3)\\ 
\text{car} & \textbf{92.9 (0.2)} &78.9 (0.1) &65.9 (0.9) &63.7 (0.6) &63.6 (0.4) &63.9 (0.3) &64.4 (0.4)\\ 
\text{deer} & \textbf{92.3 (0.1)} &82.6 (0.1) &64.3 (0.8) &62.3 (0.6) &62.8 (0.5) &63.3 (0.3) &62.4 (0.5)\\ 
\text{dog} & \textbf{93.2 (0.2)} &77.3 (0.1) &64.1 (0.7) &63.4 (0.6) &63.6 (0.4) &63.5 (0.4) &64.1 (0.3)\\ 
\text{frog} & \textbf{96.4 (0.1)} &85.8 (0.0) &67.2 (0.6) &66.4 (0.4) &65.9 (0.4) &65.8 (0.5) &65.2 (0.6)\\ 
\text{horse} & \textbf{93.6 (0.2)} &78.5 (0.1) &65.9 (0.9) &65.3 (0.4) &65.0 (0.3) &65.0 (0.5) &64.9 (0.3)\\ 
\text{ship} & \textbf{77.8 (0.4)} &72.0 (0.1) &62.8 (0.3) &61.8 (0.5) &60.9 (0.4) &61.3 (0.2) &60.9 (0.3)\\ 
\text{truck} & \textbf{89.8 (0.2)} &77.1 (0.0) &63.8 (0.3) &63.5 (0.5) &63.2 (0.6) &63.2 (0.3) &63.1 (0.5)\\ 
\hline
\end{tabular}
\end{table}
\FloatBarrier

\subsection{Additional Figures for CIFAR-10}
Similarly to the main part of the paper, we provide figures for additional eight pairs of CIFAR-10.

\begin{figure}[htb]
  \centering
\includegraphics[width=\textwidth]{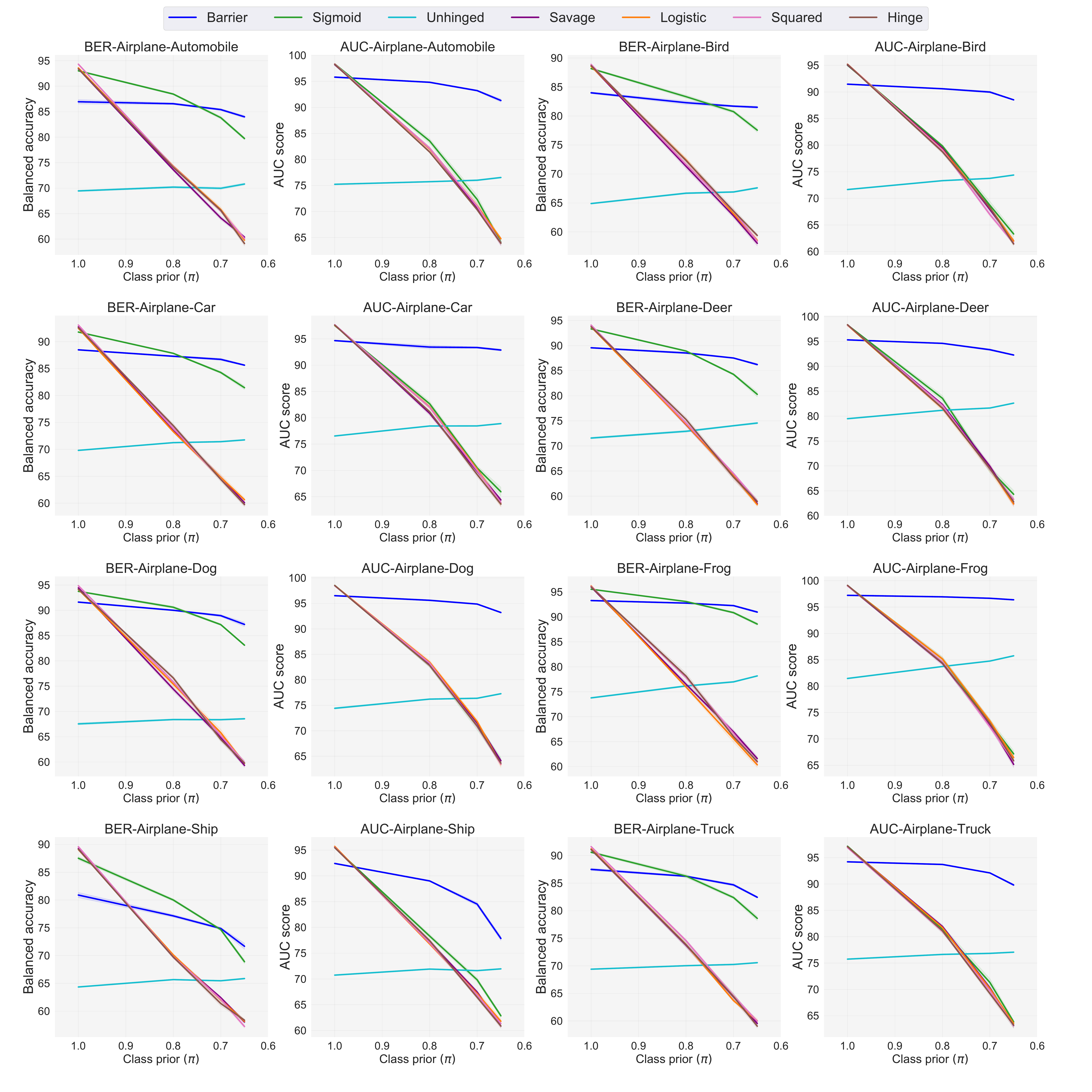}
\caption{Mean balanced accuracy (1-BER) and AUC score using convolutional neural networks (rescaled to 0-100).
The noise rate is ranged from $(\pi=1.0, \pi'=0.0), (\pi=0.8, \pi'=0.3), (\pi=0.7, \pi'=0.4), (\pi=0.65, \pi'=0.45)$.}
\end{figure}
\begin{figure}
  \centering
\includegraphics[width=\textwidth]{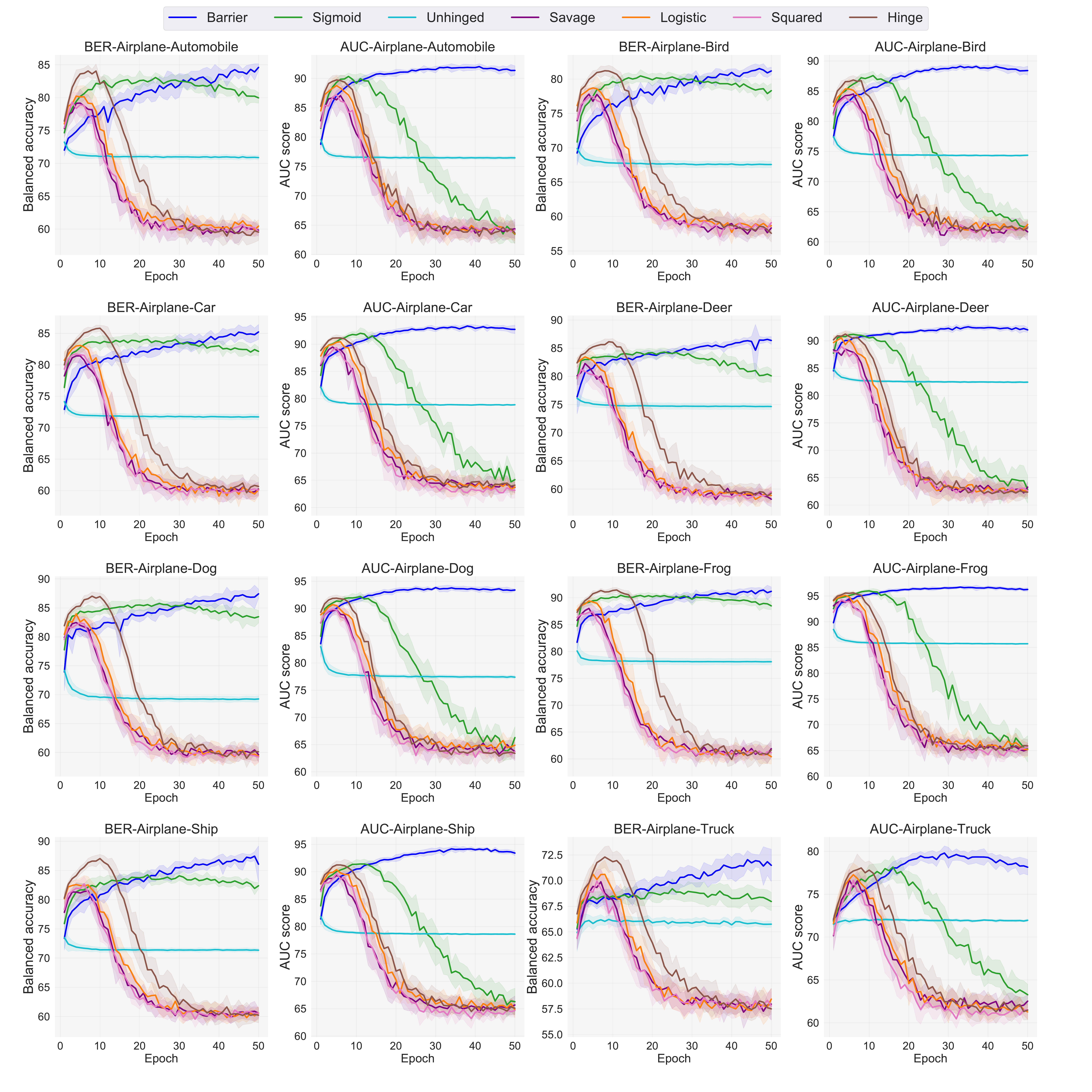}
\caption{Mean balanced accuracy (1-BER) and AUC score using convolutional neural networks (rescaled to 0-100).
The noise rate is $\pi = 0.65$ and $\pi'=0.45$. The experiments were conducted 10 times.}
\end{figure}

\end{document}